\documentclass[11pt]{article}

\usepackage[preprint]{acl}

\usepackage{times}
\usepackage{latexsym}

\usepackage[T1]{fontenc}

\usepackage[utf8]{inputenc}

\usepackage{microtype}

\usepackage{inconsolata}

\usepackage{graphicx}

\usepackage{booktabs}
\usepackage{multirow}
\usepackage{amsmath}
\usepackage{amssymb}  
\usepackage{subcaption} 
\usepackage{tabularx}  
\usepackage{array}     
\usepackage{bm}
\usepackage{enumitem}
\usepackage{textcomp}
\usepackage{colortbl}
\usepackage[table,xcdraw]{xcolor}

\usepackage{algorithm}
\usepackage{algorithmicx}
\usepackage{algpseudocode}
\floatname{algorithm}{Algorithm} 

\title{CoCoA: Collaborative Chain-of-Agents for Parametric-Retrieved Knowledge Synergy}

\author{
 \textbf{Yi Jiang\textsuperscript{1}},
 \textbf{Sendong Zhao\textsuperscript{1}\thanks{Corresponding author.}},
 \textbf{Jianbo Li\textsuperscript{1}},
 \textbf{Haochun Wang\textsuperscript{1}},
\\
 \textbf{Lizhe Zhang\textsuperscript{2}},
 \textbf{Yan Liu\textsuperscript{2}},
 \textbf{Bing Qin\textsuperscript{1}}
\\
\\
 \textsuperscript{1}Harbin Institute of Technology, China
 \\
 \textsuperscript{2}China Mobile Group Heilongjiang Co.,Ltd
\\
  \texttt{
   \{yjiang,sdzhao,jbli,hcwang,qinb\}@ir.hit.edu.cn
 }
}

\begin{document}
\maketitle

\begin{abstract}
Retrieval-Augmented Generation (RAG) enhances Large Language Models (LLMs), especially for knowledge-intensive tasks. 
Despite its advantages, current RAG methods often struggle to \textit{fully exploit knowledge during generation}. In particular, the synergy between the model’s internal parametric knowledge and external retrieved knowledge remains limited. Retrieved contents may sometimes mislead generation, while certain generated content can guide the model toward more accurate outputs. 
In this work, we propose \textbf{Co}llaborative \textbf{C}hain-\textbf{o}f-\textbf{A}gents, a framework designed to enhance explicitly synergy over both parametric and retrieved knowledge. 
Specifically, we first introduce CoCoA-zero, a multi-agent RAG framework that first performs conditional knowledge induction and then reasons answers. 
Building on this, we develop \textbf{CoCoA}, a long-chain training strategy that synthesizes extended multi-agent reasoning trajectories from CoCoA-zero to fine-tune the LLM. 
This strategy enhances the model’s capability to explicitly integrate and jointly leverage parametric and retrieved knowledge. 
Experimental results demonstrate the superiority of CoCoA in open-domain QA and multi-hop QA. Code is public\footnote{Code available at https://github.com/liunian-Jay/CoCoA .}.
\end{abstract}

\section{Introduction}
Large Language Models (LLMs)~\citep{achiam2023gpt,touvron2023LLaMA} have demonstrated strong performance across a wide range of natural language tasks. However, the knowledge they rely on is embedded in their parameters and cannot be easily updated as new information emerges~\citep{ji2023survey,he2022rethinking}. To address this limitation, the Retrieval Augmented Generation (RAG) framework introduces an external retrieval component that brings in external knowledge and integrates it into the input context of the LLMs. This design has led to notable improvements in various natural language processing applications~\citep{gao2023retrieval,lewis2020retrieval}. 
Existing research has primarily aimed to improve two aspects of RAG:  \textit{retrieving more  useful information} during retrieval and \textit{better utilizing information to guide generation} during generation. 
Despite these efforts, most retrieval-augmented language models (RALMs) still emphasize external retrieval, while paying insufficient attention to the rich internal knowledge already encoded in model parameters. 
This internal knowledge is especially valuable for open-domain question answering, where many queries are factual and often already covered during pretraining.

\begin{figure}[p!t]
    \centering
    \includegraphics[width=0.975\linewidth]{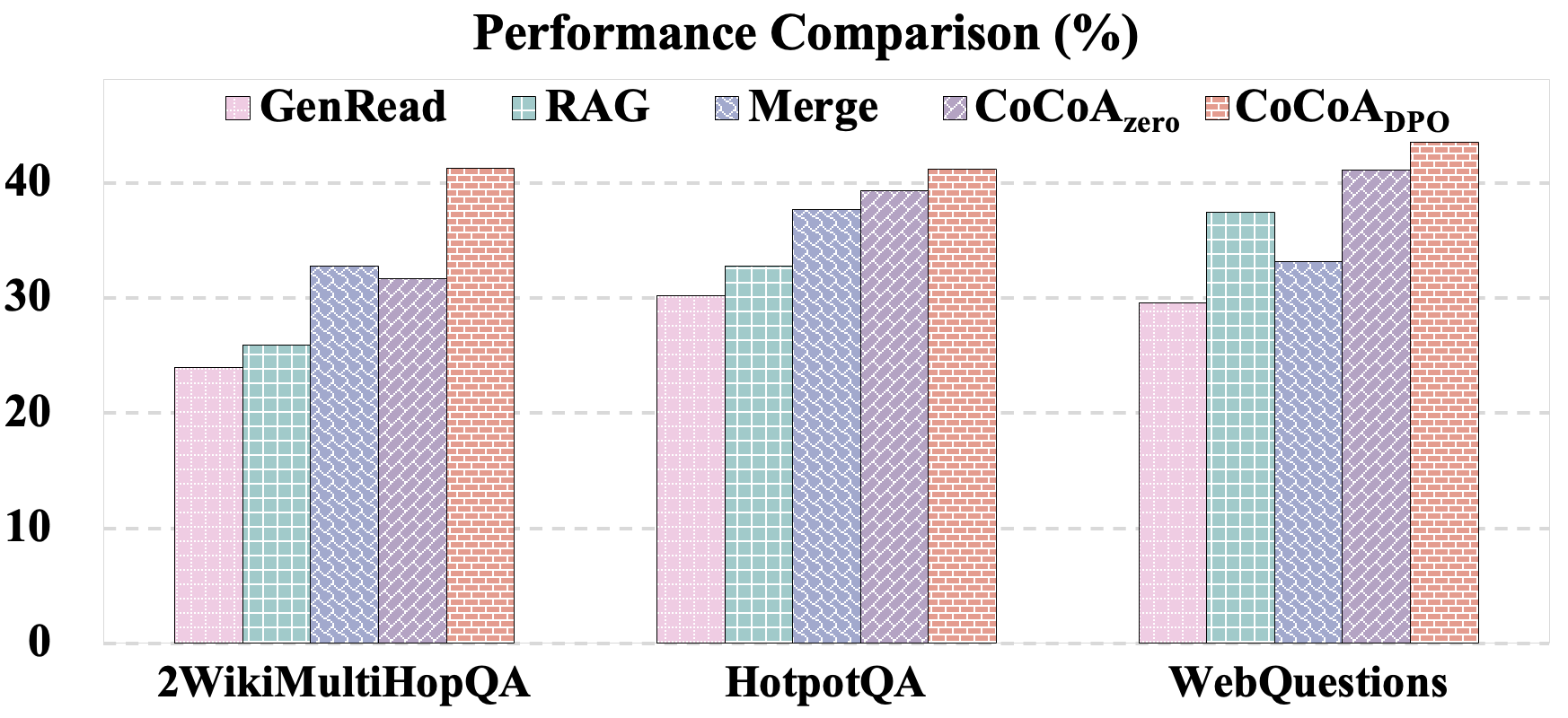}
        \caption{Evaluation on three datasets. The Merge method is a simple strategy we use to verify the collaboration of internal and external knowledge. It directly generates a passage and merges it into the retrieved passages as the context of the LLM.}
    \label{fig:challenge}
\end{figure}

\begin{figure*}[!ht]
    \centering
    \includegraphics[width=0.975\textwidth]{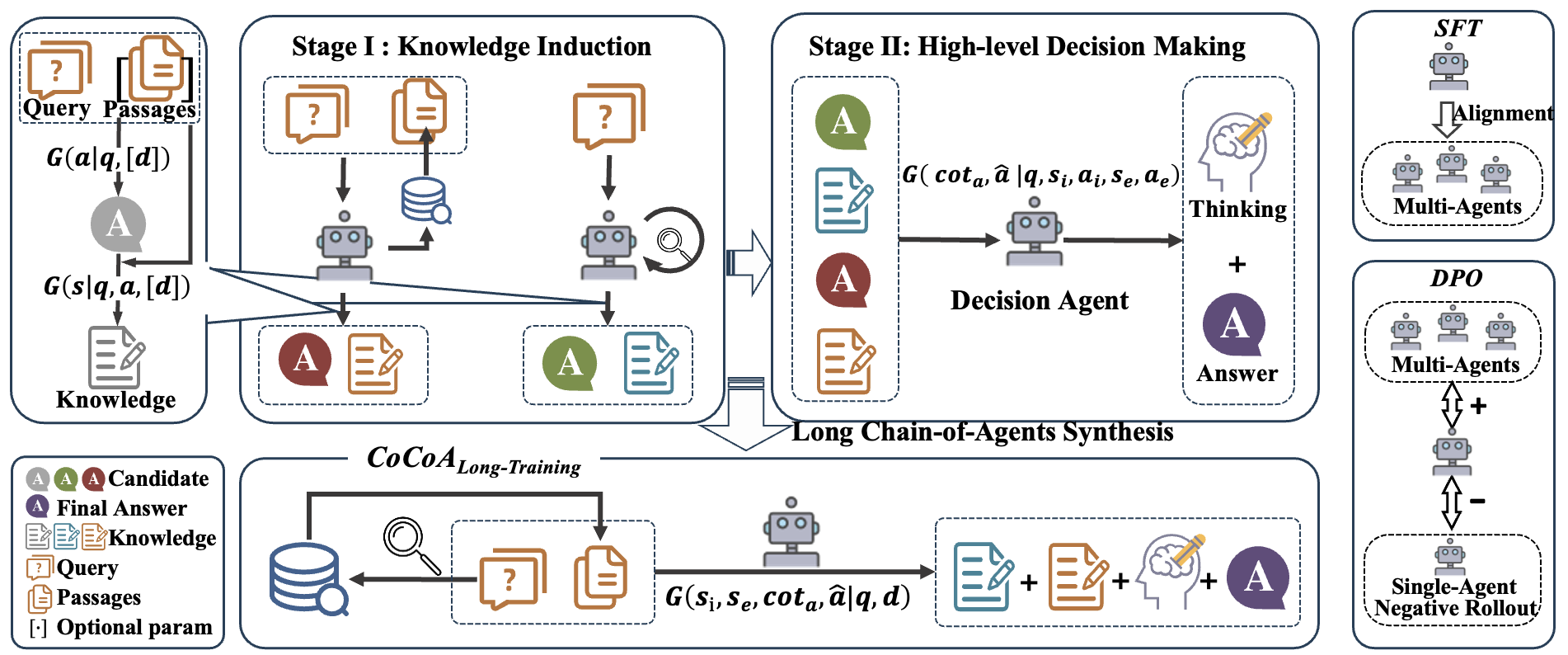}
    \caption{Illustration of the CoCoA framework. The top part is CoCoA-zero, a multi-agent collaboration framework. It integrates internal and external knowledge in a collaborative manner by first performing knowledge induction and then making decisions. The bottom part is the training strategy, which is based on CoCoA-zero and combines the trajectories of different agents into long chains to train and enhance the integration ability of the LLM.}
    \label{fig:framework}
\end{figure*}

Specifically, as the knowledge in LLM's parameter becomes richer and 
 the abilitiy of the LLM becomes stronger, sometimes answers with search information are not as good as direct answers. 
To validate the necessity of collaboratively synergizing internal (or parametric) and external (or retrieved) knowledge, we conducted experiments to compare performance. 
As shown in Fig.~\ref{fig:challenge}, across the three evaluation tasks, direct generation and GenRead~\citep{yu2022generate} (explicitly generated content) sometimes shows stronger performance. 
Also, we conduct a test experiment,``Merge’’, that explicitly integrates internal and external knowledge by combining retrieved passages with internally generated passages as the final context. 
``Merge'' often achieves better results than both direct generation and RAG approaches, demonstrating the potential of internal and external knowledge collaboration. However, its improvements are not consistent across all datasets, indicating the need for more sophisticated integration methods.


To address the above challenges, we introduce \textbf{CoCoA}, which consists of a multi-agent reasoning framework and a training strategy that combines multi-agent trajectories into long chains to enhance LLM performance. 
Specifically, we first introduce \textbf{CoCoA-zero}, which features three complementary agents: one for extracting pre-trained knowledge, one for retrieving external data, and one for reasoning over both to make optimal decisions. 
This not only enables explicit construction of decoupled internal and external knowledge, but also provides collaborative reasoning traces for the training, 
particularly the agent's ability to synthesize information and make context-aware decisions. 
Based on CoCoA-zero, we further introduce an end-to-end training strategy for \textbf{CoCoA}, which significantly improves performance on knowledge-intensive tasks by fusing the collaborative capabilities of multi-agents into one model.

Our contributions are summarized as follows:
\begin{itemize}
    \item We investigate the challenge of parametric–retrieved knowledge collaboration and introduce \textbf{CoCoA-zero}, a multi-agent reasoning framework that coordinates parametric and retrieved knowledge for improved generation. 
    \item We develop a training paradigm for CoCoA, which distills multi-agent reasoning into long-chain, enabling LLMs to better exploit internal and external knowledge. 
    \item Extensive experiments demonstrate \textbf{CoCoA}'s effectiveness, offering insights for inference-time scaling and multi-agent training on knowledge-intensive tasks.
\end{itemize}

\section{Related Works}
\subsection{Retrieval-augmented Generation}
In recent years, to address outdated knowledge and hallucination of LLM, RAG has been introduced~\citep{fan2024survey,gao2023retrieval}, and many efforts have been made in two aspects: ``\textit{how to retrieve more relevant information}’’ including retriever fine-tuning~\citep{nian2024w} and query optimization~\citep{ma2023query,wang2023query2doc,wang2024blendfilter} and ``\textit{how to better use the retrieved information}’’ including domain fine-tuning~\citep{wang2024rear,zhang2024raft,yue2025synergistic,xia2025improving} and controlled decoding strategies~\citep{shi2023trusting}. 
Our CoCoA falls into the second category: better utilization of knowledge.

\subsection{RAG Pipeline Optimization}
Pipeline optimization usually adds pre-generation processing, retrieval intent identification, or optimizes the pipeline as a whole. 
For example, \citet{glass2022re2g,kim2024re} and \citet{yu2023chain} introduce reranking and refinement steps before generation, mitigating the impact of noisy retrieved passages. 
SKR~\citep{wang2023self} and UAR~\citep{cheng2024unified} avoid unnecessary retrieval by adding retrieval intent identification processes before generation. 
SURE\cite{kim2024sure} first generates multiple candidate answers and performs conditional summary verification based on the candidate answers, allowing LLMs to focus on specific contexts. 
AstuteRAG~\citep{wang2024astute} integrates reliable information iteratively. 
However, these methods fail to effectively combine internal and external knowledge or enhance the LLM’s ability to achieve synergy between them, which can limit performance.

\subsection{RALM Enhancement}
Retrieved-Augmented Language Model(RALM) enhancement is usually achieved by adjusting the LLM to achieve effective use of the information. 
One common approach is to train the LLM itself. 
For example, RAFT~\citep{zhang2024raft} and InstructRAG~\citep{wei2024instructrag} improve robustness to noisy context via noise-resistance training.
REAR~\citep{wang2024rear} balances external and internal knowledge by training relevance-guided generation.
Self-RAG~\citep{asai2023self} trains LLMs to decide whether to perform retrieval and to improve its self-reflection capabilities. 
Another approach involves guiding the decoding~\citep{shi2023trusting, kim2024adaptive}. 
For instance, CAD~\citep{shi2023trusting} enforces trust in retrieved information by contrastive decoding
However, both approaches tend to underutilize the model’s internal knowledge, which may constrain the quality and informativeness of its responses.

\section{Methodology}
In this section, we present \textbf{CoCoA-zero} and \textbf{CoCoA}, as illustrated in Fig.~\ref{fig:framework}. 
We first describe the multi-agent framework, CoCoA-zero, followed by the long-chain training strategy for CoCoA. 
The algorithm is shown in Algorithm~\ref{alg:CoCoA}. 

\subsection{Preliminaries}
We formalize the standard RAG framework. 
Given a query $q$ and a corpus $\mathcal{D}$, the RAG system retrieves $k$ relevant passages $C = \{c_1, c_2, \cdots, c_k\} \subset \mathcal{D}$ and generates an answer $\hat{a}$ based on the combined input.
This process follows a retrieve-then-generate paradigm and can be formulated as: \begin{equation} 
\begin{split} 
C &= \mathcal{R}(q, \mathcal{D}, k),\\
\hat{a} &= G(\mathcal{P}(q, C)), 
\end{split} 
\end{equation}
where $\mathcal{R}$ is the retriever, $\mathcal{P}$ is the prompt constructor that formats $q$ and $C$, and $G$ is the generator (e.g., a LLM) that predicts the final answer $\hat{a}$.

\subsection{Two-stage Framework: CoCoA-zero}
\label{method:CoCoA-zero}
In this section, we present our multi-agent RAG framework, CoCoA-zero, which also functions as the data synthesis pipeline for CoCoA. 
 As shown in Fig.~\ref{fig:framework}, Stage \MakeUppercase{\romannumeral 1} (\textsection~\ref{sec:stage1}) employs two specialized agents to induce knowledge from parameters and retrieval, while Stage \MakeUppercase{\romannumeral 2} (\textsection~\ref{sec:stage2}) introduces a agent to synthesize them for high-level decision-making. 

\subsubsection{Stage \MakeUppercase{\romannumeral 1}: Knowledge Induction.}
\label{sec:stage1}

It is challenging to extract implicit knowledge solely from the model’s internal knowledge or retrieved passages.
Inspired by GenRead~\citep{yu2022generate} and SURE~\citep{kim2024sure}, we design two dedicated agents for knowledge induction. Each agent first generates an answer and then summarizes knowledge based on that answer.

\paragraph{\textit{Internal Knowledge Induction Agent.}}
Directly allowing the model to explicitly generate its own internal knowledge is difficult to control and will inevitably result in sparse or inconsistent knowledge being generated. 
Following SURE~\citep{kim2024sure}, we introduce conditional induction. 
Specifically, the Internal Knowledge Agent samples a candidate \( a_{\text{in}} \) from the LLM based on the question: 
\begin{equation}
    a_{\text{in}} = G( \mathcal{P}(q) )
\end{equation}
Next, we prompt the LLM to generate a knowledge passage \( s_{\text{in}} \) conditioned on $q$ and \( a_{\text{in}} \), which reflects the model’s internal understanding:  
\begin{equation}
    s_{\text{in}} = G( \mathcal{P}(q, a_{\text{in}})). 
\end{equation} 

\paragraph{\textit{External Knowledge Induction Agent.}}
For retrieved passages, the External Knowledge Agent follows a similar procedure. 
Specially, it first retrieve some passages $C = \{c_1, c_2,\cdots , c_k\}$ from the corpus $\mathcal{D}$. Conditioned on both \( q \) and \( C \), it produces a second candidate \( a_{\text{ex}} \):
\begin{equation}
    a_{\text{ex}} = G(\mathcal{P}(q, C))
\end{equation}
Then, conditioned on \( q \), \( a_{\text{ex}} \) and \( C \), the agent induces the external knowledge passage \( s_{\text{ex}} \), :
\begin{equation}
    s_{\text{ex}} = G(\mathcal{P}(q, a_{\text{ex}}, C)). 
\end{equation}

The conditional knowledge induction framework thus: (1) renders implicit knowledge explicit and controllable; (2) serves as a secondary verification of responses; and (3) establishes a solid foundation for high-level decision-making in the next stage.

\subsubsection{Stage \MakeUppercase{\romannumeral 2}: High-level Decision Making.}
\label{sec:stage2}
Building on the candidate answers and inductive knowledge obtained in Stage I, the second stage leverages the LLM’s reasoning ability to perform high-level decision making.

\paragraph{\textit{Decision-Making Agent.}}
The Decision-Making Agent adopts COT~\citep{wei2022chain} reasoning over the internal and external candidate answers and their corresponding knowledge. 
It will be prompted with all five components (questions, internal and external candidate answers and their corresponding inductive knowledge) and generate the final answer \( \hat{a} \) through COT.
\begin{equation}
    cot_{\text{a}},\hat{a} = G(\mathcal{P}_{\text{cot}}( q, s_{\text{in}},a_{\text{in}}, s_{\text{ex}}, a_{\text{ex}}))
\end{equation}
Here, \(cot_{\text{a}}\) denotes the reasoning path 
that drives explicitly decision-making and guides final answer generation.

The model thereby functions as a high-level aggregator, reinforcing potentially consistent beliefs and resolving potential conflicts between internal beliefs and retrieved evidence. 
By explicitly modeling and comparing knowledge before committing to an answer, our framework improves the transparency and robustness of the decision process.

\subsection{Collaborative Chain-of-Agents Training}
Although multi-agent collaboration for internal and external knowledge coordination is simple and effective, how to achieve global optimization across multi-agents remains non-trivial. 

To this end, we propose the Collaborative Chain-of-Agents training strategy, which aims to optimize multi-agent collaboration end to end by supervising the LLM on long-form reasoning trajectories. 
These trajectories are synthesized from the multi-agent pipeline CoCoA-zero (\textsection~\ref{method:CoCoA-zero}) and reflect the full reasoning process that integrates both parametric and retrieved knowledge.


\subsubsection{Supervised Fine-Tuning.}
The CoCoA-zero framework is designed to: (1) control the direction of knowledge generation via conditional induction, (2) decouple internal and external knowledge through parallel reasoning, and (3) integrate both sources via Chain-of-Thought decision making. 

To supervise the model to achieve explicit and collaborative knowledge integration, we synthesize training samples by concatenating the intermediate results produced by CoCoA-zero into a single long-form response. 
Specifically, given a question \(q\) and a set of retrieved documents \(C\), we integrate the intermediate results from the CoCoA-zero (i.e., internal induction \(s_{\text{in}}\), external induction \(s_{\text{ex}}\), the CoT reasoning trace \(cot_{\text{a}}\) during integration and the final answer \( \hat{a} \) ) into a long response \(y\) and promote the evolution of model capabilities through the following supervision objectives:
\begin{equation}
\label{eq:sft}
    \mathcal{L}_{\text{SFT}} = -\mathbb{E}_{(x, y) \sim \mathcal{D}} \big[ \log P_{\theta}(s_{\text{in}}, s_{\text{ex}}, cot_{\text{a}}, \hat{a} \mid q, d)\big].
\end{equation}

This training explicitly exposes the model to long collaborative samples, where the target outputs are synthesized based on  CoCoA-zero. Through end-to-end training, multiple agents can influence and enhance each other's capabilities. Moreover, the noise introduced by intermediate agents becomes negligible, as it contributes to the overall robustness of the training process.

\begin{algorithm}[!h]
\caption{CoCoA: Example of one sample}
\label{alg:CoCoA}
\textbf{Input:} Query $q$, corpus $\mathcal{D}$, hyperparameters $k$ \\
\textbf{Output:} Final answer $\hat{a}$ or training sample $y$
\begin{algorithmic}[1]
    \State \textit{\textbf{CoCoA-zero}}: 
    \begin{algorithmic}[1]
        \State $a_{\text{in}} \gets G_{\text{in}}(\mathcal{P}(q))$ \Comment{Internal candidate}
        \State $s_{\text{in}} \gets G_{\text{in}}(\mathcal{P}(q, a_{\text{in}}))$ \Comment{Internal knowledge induction}
        \State $C \gets \mathcal{R}(q, \mathcal{D}, K)$ \Comment{Top-$K$ retrieval} 
        \State $a_{\text{ex}} \gets G_{\text{ex}}(\mathcal{P}(q, C))$ \Comment{External candidate}
        \State $s_{\text{ex}} \gets G_{\text{ex}}(\mathcal{P}(q, a_{\text{ex}}, C))$ \Comment{External knowledge induction}
        \State $(cot_{\text{a}}, \hat{a}) \gets G_{\text{dm}}(\mathcal{P}(q, s_{\text{in}}, s_{\text{ex}}, a_{\text{in}}, a_{\text{ex}}))$ \Comment{Decision making}
    \end{algorithmic}
\end{algorithmic}

\begin{algorithmic}[1]
    \setcounter{ALG@line}{1}
    \If{Supervised Fine-tuning}
        \State $y \gets (s_{\text{in}} \oplus s_{\text{ex}} \oplus cot_{\text{a}} \oplus  \hat{a})$ \Comment{CoCoA Target}
        \State Update model with $\mathcal{L}_{\text{SFT}}$  in Eq.~\ref{eq:sft}.
    \EndIf
    \If{DPO Training}
        \State $y^{-} \gets G( \mathcal{P}_{\text{ZS}}(q, C))$ 
        \State $y^{+} \gets (s_{\text{in}} \oplus s_{\text{ex}} \oplus cot_{\text{a}} \oplus \hat{a})$
        \State Update model with $\mathcal{L}_{\text{DPO}}$ in Eq.~\ref{eq:dpo}
    \EndIf
    \State \Return $\hat{a}$ or the trained model CoCoA
\end{algorithmic}
\end{algorithm}

\subsubsection{Preference optimization.}
\label{sec:dpo}
To better align the model with collaborative multi-agent behavior, we apply DPO~\citep{rafailov2023direct} training using positive samples from CoCoA-zero and negative ones from a zero-shot single-agent variant. 
The key insight is that single-agent responses often show biased or fragmented reasoning, such as over-relying on retrieval or ignoring internal signals. 
Note that this can be seen as a special case of SFT using both positive and negative samples, rather than reinforcement learning. 
Each training instance includes a context \(x = (q, d)\), a preferred response \(y^{+} = (s_{\text{int}} \oplus s_{\text{ext}} \oplus t \oplus  \hat{a})\) from the CoCo-zero, and a rejected response \(y^{-}\) from the single-agent variant.  
It encourages the model to prefer \( y^{+} \) over \( y^{-} \)  by optimizing:
\begin{equation}
\label{eq:dpo}
\begin{aligned}
\mathcal{L}_{\text{DPO}}&(\pi_\theta) = 
 - \mathbb{E}_{(x, y^+, y^-) \sim \mathcal{D}} \big[ \log \sigma\big(\beta \cdot \log \pi_\theta(y^+|x) \\
& - \beta \cdot \log \pi_\theta(y^-|x) \big) + \alpha \cdot \big( -\log \pi_\theta(y^+|x) \big) \big]
\end{aligned}
\end{equation}
where \( \pi_\theta(y|x) \) denotes the unnormalized log-probability of response \( y \) under the model \( \theta \). 

The CoCoA traing thus bridges symbolic multi-agent collaboration and end-to-end generation, enabling the model to internalize structured reasoning through supervision.


\section{Experiments}
In this section, we report our experiments results, and provide a analysis of them. More supplements are in the Appendix.

\subsection{Implementation Details}
\noindent\textbf{Training Data}
We sample subsets from the training sets of HotpotQA~\citep{yang2018hotpotqa}, 2WikiMultiHopQA~\citep{ho2020constructing} and WebQuestions~\citep{berant2013semantic}, then synthesize data using the CoCoA-zero and filter them based on gold answers.  
This results in 6.8k filtered samples for SFT. 
For DPO, we select 1151 samples, which are the ones that are answered incorrectly by zero-shot but correctly by the CoCoA-zero framework. 
For each sample, we gather 5 relevant passages using CONTRIEVER~\citep{izacard2021unsupervised}.

\noindent\textbf{Training Details}
We fine-tune LLaMA3.1-8B with LoRA (r=16, $\alpha$=16, dropout=0.05). 
During SFT, we train for 5 epochs with a learning rate of 3e-5. For DPO, we used $\beta$=0.2 and $\alpha$=0.2 (RPO), with a learning rate of 5e-6. 
All experiments are conducted on a single A100 GPU.

\noindent\textbf{Inference Details}
During inference, we use Contriever~\citep{izacard2021unsupervised} as the retriever and set k to 5. 
For all datasets, we use 21M English Wikipedia~\citep{karpukhin2020dense} dump as the source passages for the retrieval. 
Prompts for the experiments can be found in Appendix~F.

\subsection{Datasets and Evaluation Metrics}
\textbf{Evaluation Datasets} 
To evaluate the effectiveness of CoCoA, we conduct experiments on open-domain QA task: WebQuestions~\citep{berant2013semantic}, and TriviaQA~\citep{joshi2017triviaqa}, as well as multi-hop QA task: HotpotQA~\citep{yang2018hotpotqa} and 2WikiMultiHopQA~\citep{ho2020constructing}.
Details are provided in Appendix~\ref{sec:appendix-dataset}.


\noindent \textbf{Evaluation Metrics} 
We report both Exact Match (EM) and F1 scores. Following~\citet{asai2023self,mallen2022not}, we adopt a non-strict \textbf{EM} metric that deems a prediction correct if it contains the gold answer. F1 measures token-level overlap between the predicted and gold answers. In our setting, longer responses often yield higher \textbf{EM} scores, 
but may reduce \textbf{F1}.
Thus, considering both metrics provides a more balanced evaluation.

\begin{table*}[!ht]
\centering
\resizebox{0.975\linewidth}{!}{%
\begin{tabular}{lcccccccccccc}
\toprule
\multicolumn{1}{c}{}     & \multicolumn{3}{c}{2WikiMQA} & \multicolumn{3}{c}{HotpotQA} & \multicolumn{3}{c}{WebQuestions} & \multicolumn{3}{c}{TriviaQA$^{\bm\ddagger}$} \\
\multicolumn{1}{c}{\multirow{-2}{*}{Method}} 
& EM     & F1  & Avg            
& EM     & F1  & Avg     
& EM     & F1  & Avg      
& EM     & F1  & Avg   \\
\hline

\multicolumn{13}{c}{\cellcolor[HTML]{EFEFEF}Llama-3.1-Instruct Train-free \& w/o retrieval}\\

Llama-3.1-8B 
& \underline{27.60}  & \underline{28.35} & \underline{27.98}  
& 24.00  & 27.09 & 25.54 
& \underline{40.11}  & \textbf{39.98} & \underline{40.04}
& 62.87         & 64.17 & 63.52 \\

8B+COT                   
& 23.80            & 26.55     & 25.28       
& 26.20         & 32.26     & 29.23   
& 38.04           & 39.43   & 38.73       
& 64.90         & 66.98     & 65.94   \\

8B+GenRead                
& 24.00         & 23.92   & 23.96      
& 29.20         & 31.15   & 30.18     
& 29.53         & 29.67   & 29.60   
& 54.12         & 54.29   & 54.21   \\

\textit{Llama-3.1-70B}
& \textit{33.80}    & \textit{33.43}      & \textit{33.62}
& \textit{37.00}    & \textit{37.89}      & \textit{37.45} 
& \textit{44.83}    & \textit{43.92}      & \textit{44.38}  
& \textit{77.89}    & \textit{78.93}     & \textit{78.81}    \\

\multicolumn{13}{c}{\cellcolor[HTML]{EFEFEF}Llama-3.1-Instruct Train-free \& w/ retrieval}\\

8B+StandardRAG                         
& 26.80            & 25.07     & 25.94       
& 31.40         & 34.16   & 32.78     
& 37.65           & 37.32 & 37.49             
& \underline{66.83}         & 67.16   & \underline{66.99}     \\

8B+COT                   
& 22.40            & 25.25     &  23.83   
& 32.40         & 38.71     & 35.55    
& 35.73           & 36.17 & 35.95   
& 65.85         & \underline{67.54}     & 66.69    \\

8B+CON                   
& 19.00            & 21.32     & 20.16        
& \underline{32.80} & \underline{38.67}    & \underline{35.73} 
& 34.40           & 38.05     & 36.22       
& 65.64           & 66.82      & 66.23  \\

8B+SURE                  
& 18.40   & 21.32  & 19.86       
& 32.00   & 37.26  & 34.63   
& 32.48   & 39.01  & 35.75   
& 63.14   & 62.91  & 63.02   \\

CoCoA-zero-8B      
& \textbf{31.40}    & \textbf{31.92}  & \textbf{31.66}  
& \textbf{37.40}    & \textbf{41.20}  & \textbf{39.30}  
& \textbf{43.11}    & \underline{39.13} & \textbf{41.12}  
& \textbf{70.73}    & \textbf{69.99}   & \textbf{70.36}  \\

\textit{Llama-3.1-70B}                        
& \textit{22.00}            & \textit{23.12}   & \textit{22.56}      
& \textit{35.20 }        & \textit{38.03}    & \textit{36.61} 
& \textit{39.76}           & \textit{39.05}   & \textit{39.41} 
& \textit{70.97}         & \textit{71.44 }    & \textit{71.20}  \\

\midrule
\multicolumn{13}{c}{\cellcolor[HTML]{EFEFEF}RALM w/ retrieval \& w/ Training}\\

Self-RAG 7B                                  
& 37.40         & 17.93     & 27.66        
& 33.40         & 20.57     & 26.99      
& 44.64         & 25.75     & 35.19      
& 66.30         & 37.27     & 51.78    \\

Self-RAG 13B                                 
& 38.80         & 22.61       & 30.71      
& 35.40         & 21.64      & 28.52 
& \textbf{45.87}  & 25.31     & 35.59      
& 68.74         & 38.22   & 53.48   \\

DeepSeek-R1-8B                 
& 36.80         & 25.79     & 31.30        
& 35.00         & 32.66    & 33.83     
& 44.34         & 31.87    & 38.11   
& 65.62         & 58.07    & 61.84   \\

InstructRAG-8B
&36.40 	&\underline{39.40} 	&37.90 
& $-$ & $-$ & $-$
& $-$ & $-$ & $-$
&\underline{70.90} 	&65.40 	&68.15 \\

CoCoA-SFT-8B                                  
& \underline{41.00}            & 36.87     & \underline{38.94} 
& \textbf{39.40}         & \textbf{46.31}    & \textbf{42.86} 
& 42.96           & \underline{41.32}    & \underline{42.14} 
& 70.72        & \underline{70.39}   & \underline{70.55} \\

CoCoA-DPO-8B                                  
& \textbf{42.00}            & \textbf{40.58}        & \textbf{41.29}     
& \underline{39.00}         & \underline{43.39}    & \underline{41.20}     
& \underline{44.83}       & \textbf{42.21}    & \textbf{43.52} 
& \textbf{71.52}         & \textbf{70.42}       & \textbf{70.97}  \\

\bottomrule
\end{tabular}
}
\caption{EM/F1 of different methods experimented on four datasets. The best and second best scores are highlighted in \textbf{bold} and \underline{underlined}, respectively. \textit{Italics} mark a boundary, not for comparison. $^{\bm\ddagger}$ represents the Out-of-Distribution evaluation dataset.}
\label{tab:main-result}
\end{table*}

\subsection{Baselines}
We selected several of the most representative methods for comparison. 
(1) StandardRAG, which is the classic ``retrieve-then-read’’ paradigm. 
(2) Chain-Of-Thought~\citep{wei2022chain}: Uses CoT prompting to reason before answering. 
(3) Chain-Of-Note~\citep{yu2023chain}: Refines the retrieved passages prior to answering. 
(4) GenRead~\citep{yu2022generate}: Generates context to answer. 
(5) SURE~\citep{kim2024sure}: Conditional summarization followed by multiple validation. 
(6) Self-RAG~\citep{asai2023self}: Employs adaptive retrieval and self-reflection to decide when and how to use external context. 
(7) DeepSeek-R1-Distill-8B~\citep{guo2025deepseek}: A distilled LLaMA-8B model released by DeepSeek-R1, trained on reasoning data. 
(8) InstructRAG~\citep{wei2024instructrag}: Denoising training using self-synthesized data. 
All retrieval-based methods use top-5 passages. 
Details experimental settings are shown in the Appendix~\ref{sec:appendix-detail}.

\subsection{Main Results}
Experimental results are presented in Table~\ref{tab:main-result}, and we summarize the key findings as follows:

\textbf{(1) Retrieval vs. non-retrieval.}
On WikiMQA and WebQuestions, direct generation performs better, while retrieval methods excel on other tasks.
This demonstrates that retrieved knowledge and parametric knowledge each have their own strengths and weaknesses in different scenarios.

\textbf{(2) RAG without training.}
The improvements of some process optimization methods are decreasing compared to standardRAG. 
We speculate that this is because current LLMs are becoming more powerful enough to make good use of external knowledge. 
CoCoA-zero improves the average EM and F1 of all tasks by \textbf{4.99\%} and \textbf{4.64\%} respectively, while other train-free methods show little effect. 
These results suggest that current QA tasks should place greater emphasis on leveraging the model’s rich internal knowledge. 

\textbf{(3) Superiority and Generalization of CoCoA.}
Our CoCoA methods achieve state-of-the-art performance across almost all datasets. 
In particular, CoCoA improves the EM and F1 of 2WikiMultiHopQA tasks by \textbf{15.2\%} and \textbf{15.51\%} respectively.
Moreover, despite being trained with limited data, CoCoA also performed well on out-of-distribution dataset, demonstrating its robustness. 

\textbf{(4) Reasoning Distillation vs. CoCoA Training.} 
DeepSeek-R1-8B, trained on distilled reasoning data, outperforms the undistilled StandardRAG. 
CoCoA, distilled with multi-agent self-synthesis on knowledge-intensive tasks, further surpasses DeepSeek-R1-8B. 
We speculate this is because logical reasoning and knowledge-intensive tasks differ, and CoCoA can better leverage knowledge. 
This suggests that explicitly leveraging key internal and external knowledge can be more effective than chain-of-thought reasoning. 

\textbf{(5) Effect of DPO.} 
Comparing our SFT and DPO variants, DPO training yields improvements across several datasets. 
This suggests that contrastive preference learning can help the model better align to the collaborative responses of multi-agents. 
However, it may also lead to performance degradation due to the quality of training data. 

\subsection{Ablation Study I: Different Agents}
To better understand the contribution of each module in CoCoA-zero, we conduct an ablation study by selectively removing internal/external induction and the reasoning. 

As shown in Table~\ref{tab:ablation-study-prompt}, removing internal induction significantly degrades performance, especially by \textbf{8.4\%} on 2WikiMQA. This shows the importance of leveraging parameterized knowledge in scenarios such as 2WikiMultiHopQA where the LLM itself can answer well. Similarly, excluding external induction also leads to a noticeable performance drop across all datasets, highlighting the complementary role of retrieved knowledge. 
Moreover, disabling the reasoning mechanism in decision making results in a consistent decrease, suggesting that reasoning over both knowledge contributes to deeper understanding.

To further validate the effectiveness of multi-agent collaboration, we introduce a zero-shot variant using a single agent. Its performance is much lower than CoCoA-zero, which confirms the necessity of using multi-agent roles to coordinate between internal and external knowledge. 

Overall, these results confirm the effectiveness of our multi-agent collaboration design, where each component plays a non-trivial role in achieving optimal performance.

\begin{table}[!t]
\centering
\resizebox{0.999\linewidth}{!}{%
\setlength{\tabcolsep}{2.2pt}  
\begin{tabular}{llll}
\toprule
\multicolumn{1}{c}{\multirow{1}{*}{Method}} & \multicolumn{1}{c}{2WikiMQA} & \multicolumn{1}{c}{HotpotQA} & \multicolumn{1}{c}{WebQuestions} \\

\midrule
CoCoA-zero             & \textbf{31.66}      & \textbf{39.30}      & \textbf{41.12}   \\

$w/o$ Internal            & 23.26 {($\downarrow$ 8.40)}    & 36.56 {($\downarrow$ 2.74)}    & 39.10 {($\downarrow$ 2.02)}    \\

$w/o$ External             & 28.97 {($\downarrow$ 2.69)}     & 30.96 {($\downarrow$ 8.34)}     & 38.97 {($\downarrow$ 2.15)}    \\

$w/o$ Think             & 30.38 {($\downarrow$ 1.28)}     & 37.17 {($\downarrow$ 2.13)}     & 39.75 {($\downarrow$ 1.37)}   \\

\midrule

\textit{Zero-Shot}       & 18.55 {($\downarrow$ 13.11)}    & 35.01 {($\downarrow$ 4.29)}     & 35.38 {($\downarrow$ 5.74)}  \\

\textit{Standard}     & 25.94 {($\downarrow$ 5.72)}     & 32.78 {($\downarrow$ 6.52)}     & 37.49 {($\downarrow$ 3.63)}   \\
\bottomrule
\end{tabular}
}
\caption{Ablation study on knowledge induction and decision-making. The zero-shot variant (\textsection~\ref{sec:dpo}) is also included. 
We adopt the EM/F1 average as metric.
}
\label{tab:ablation-study-prompt}
\end{table}

\subsection{What is the Impact of Internal Conditional Induction?}
\begin{figure}[!hbt]
    \centering
    \includegraphics[width=0.9\linewidth]{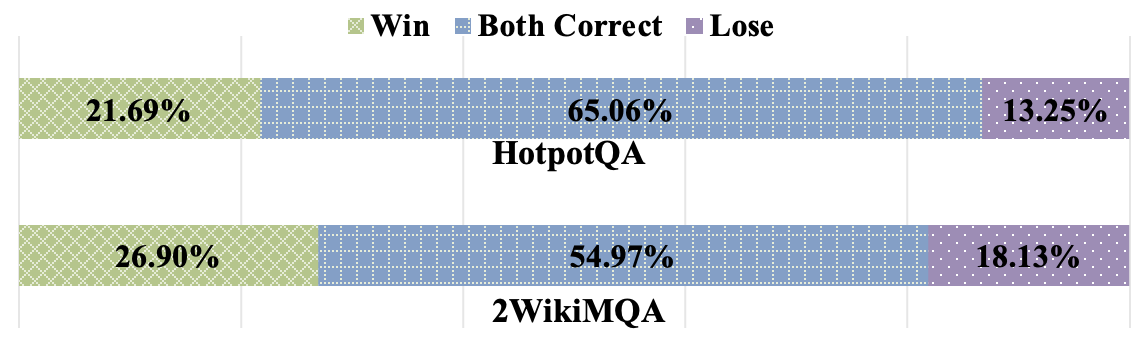}
    \caption{Illustration of Conditional Induction, with EM as the metric.}
    \label{fig:ConInternal}
\end{figure}

To explore whether internal knowledge generation can introduce harm, we conducted a qualitative analysis of our sample. Specifically, we analyzed (1) answers generated using internal knowledge and (2) answers generated directly, as shown in Figure~\ref{fig:ConInternal}. We found that internal knowledge generation had some failures, but successes far outweighed failures.
This suggests that conditional internal knowledge induction is a knowledge induction method and has a certain secondary validation effect, but it inevitably introduces some adverse effects such as hallucinations, which require more fine-grained control in the future.

\begin{figure}[!hbt]
    \centering
    \includegraphics[width=0.925\linewidth]{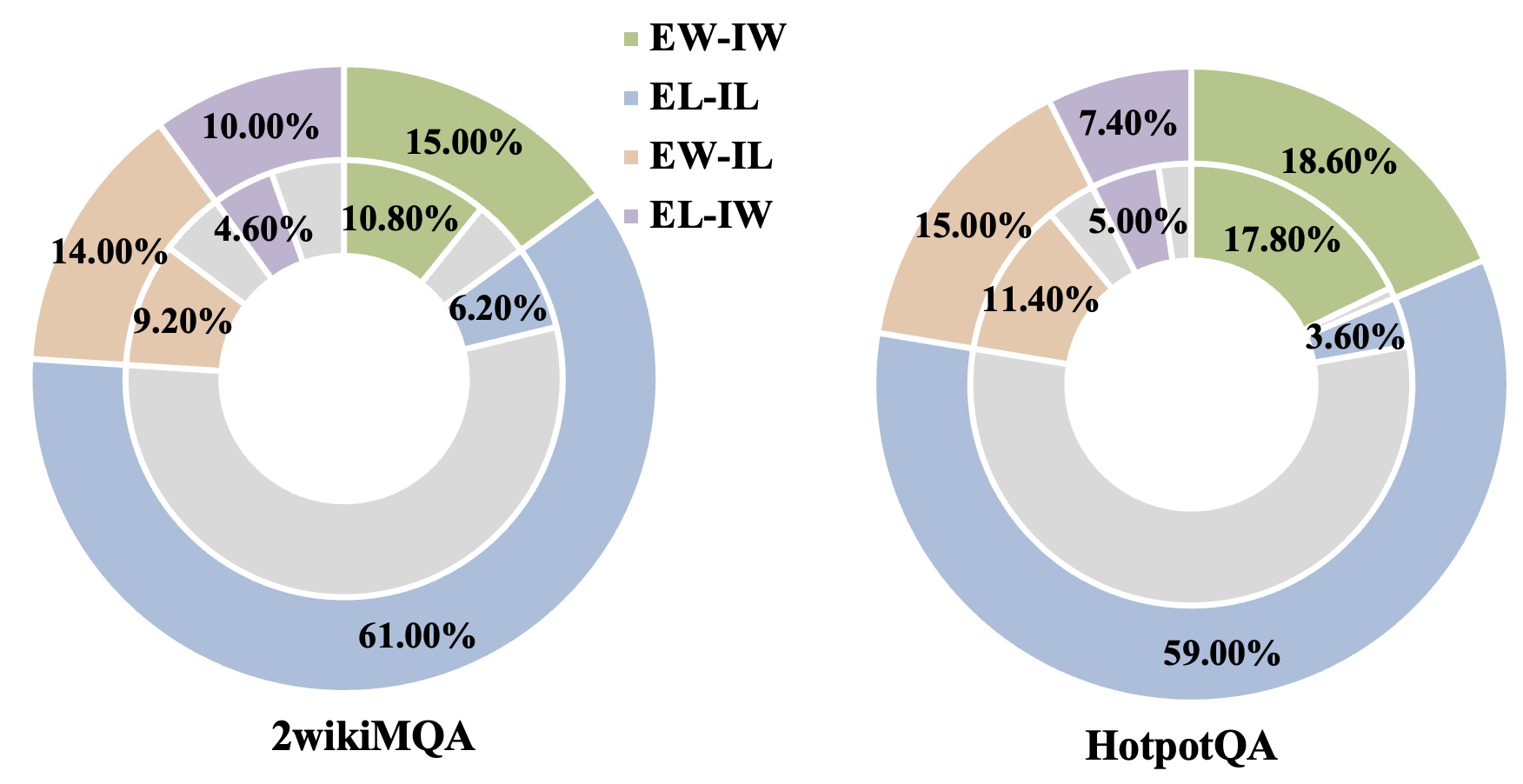}
    \caption{
    Qualitative Case Analysis (EM Metric): EW-IW (both win), EL-IL (both lose), EW-IL (external wins), EL-IW (internal wins). The outer loop is the direct answer of the internal and external agents, and the inner loop is CoCoA-zero. 
    }
    \label{fig:win-lose-tie}
\end{figure}
\subsection{Qualitative Case Study}
To analyze the mechanism of CoCoA-zero's effectiveness, we conducted a detailed qualitative analysis of the cases. 
The outer loop represents the proportion of responses directly generated by internal and external responses. 
We find that both achieved success and failed responses, demonstrating a contest of knowledge. 
The inner loop represents CoCoA-zero responses. 
We find that CoCoA-zero effectively handles cases where both responses are correct, achieving coverage rates of 57\%+ on 2WikiMQA and 73\%+ on HotpotQA, surpassing the coverage of individual responses. 
This is inconsistent with the conclusion that some work LLM prefers internally generated content\citep{tan-etal-2024-blinded}. 
We speculate that conditional induction may alleviate this bias.

Furthermore, we find that even when both responses failed, CoCoA-zero also achieved some success. 
This stems from the mutual enhancement between knowledge induction and high-level decision-making, which jointly enable the LLM to better activate its own capabilities.

\begin{table}[!h]
\centering
\resizebox{0.975\linewidth}{!}{%
\setlength{\tabcolsep}{4pt}  
\begin{tabular}{lcccc}
\toprule
\multicolumn{1}{c}{Method}        & 2Wiki   & HotpotQA   & WebQ  & Average     \\
\midrule
$\text{Long-DPO}_{8B} $           & \textbf{41.29} & \underline{41.20}  & \textbf{43.52}  & \textbf{42.00}   \\
$\text{Long-SFT}_{8B} $           & \underline{38.94}      & \textbf{42.86} & \underline{42.14}       & \underline{41.31}  \\
$\text{Short-SFT}_{8B}$           & 33.91      & 40.04      & 40.13           & 38.03\\
$\text{Short-SFT}_{8B \times 3}$  & 28.31      & 40.58      & 39.84           & 36.24 \\
\bottomrule

\end{tabular}%
}
\caption{Ablation study of the training strategy for CoCoA. For fairness, Avg(EM,F1) is used as the metric.}
\label{tab:ablation-study-train}
\end{table}
\subsection{Ablation Study II: Training Strategies}
\label{Ab:Training}
To evaluate the effectiveness of our training strategy for CoCoA, we conduct an ablation study comparing different training configurations on the LLaMA3.1-8B model. As shown in Table~\ref{tab:ablation-study-train}, $\textit{Long-DPO}_{8B}$ achieves the best overall performance, confirming the benefit of aligning long-form outputs via long-chain optimization. 

The $\textit{Short-SFT}_{8B \times 3}$ variant, where each task segment is trained on a separate model, shows clear degradation in performance, especially on 2WikiMultiHopQA. 
This indicates that separating induction and reasoning capabilities into isolated modules weakens the model’s ability to holistically integrate information across steps. 
The $\textit{Short-SFT}_{8B}$ variant, which combines three instruction capabilities into a single model but retains short-form generation, performs better than $\textit{Short-SFT}_{8B \times 3}$ but still falls behind our approaches. This shows that simply merging instructions is slightly less performant than our long chain consolidation. 

Our training strategy for CoCoA, represented by $\textit{Long-DPO}_{8B}$ and $\textit{Long-SFT}_{8B}$ variants, explicitly modeled multi-agent collaboration as a unified long-form output. The superior performance of these models underscores the advantage of training models to generate cohesive and contextually rich responses rather than fragmented predictions. This, to a certain extent, provides new perspectives for the expansion of knowledge-intensive long chains.

\subsection{Training Generalization to Non-QA Tasks}
\label{sec:task}
\begin{figure}[!hbt]
    \centering
    \includegraphics[width=0.925\linewidth]{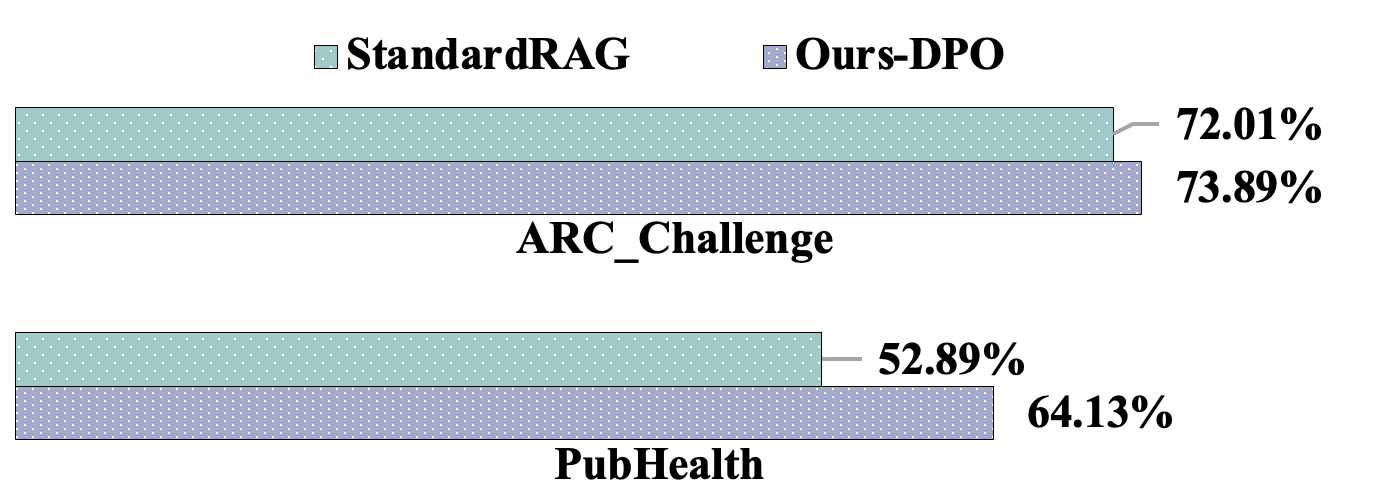}
    \caption{Illustration of accuracy changes when transferring to non-QA tasks, with accuracy as the metric.}
    \label{fig:other-task}
\end{figure}

To further evaluate the generalization of CoCoA, we test it on fact verification and multiple-choice tasks. 
As shown in Figure ~\ref{fig:other-task}, our training did not reduce the performance of these tasks compared to standard RAG. 
In fact, in some cases, we even observed a slight improvement. 
One explanation is that our training strategy encourages collaborative output that leverages the capabilities of the LLM, rather than injecting knowledge directly, and thus possesses a certain degree of universality. 

\subsection{When the Number of K Changes}
\begin{figure}[!t]
    \centering
    \includegraphics[width=0.999\linewidth]{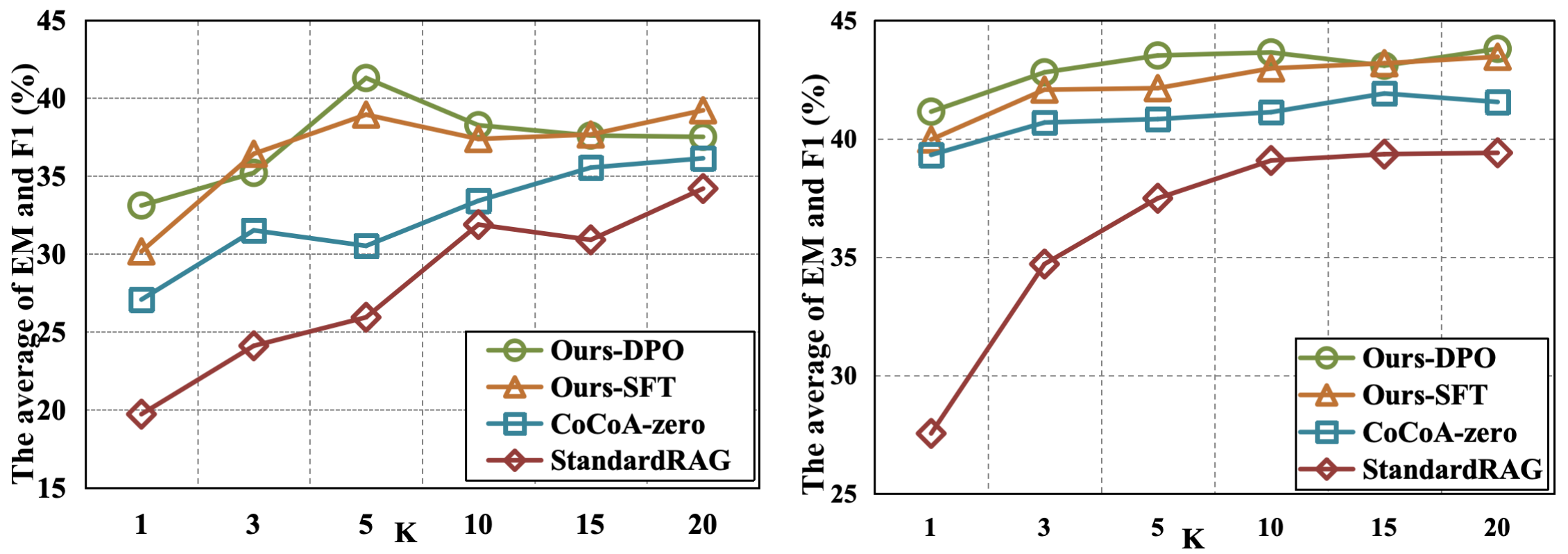}
    \caption{Performance varies with passage num: 2WikiMQA (left), WebQuestions (right).}
    \label{fig:k}
\end{figure}

In order to better explore the robustness of our CoCoA with respect to the number of documents, we set $K$ to vary in the interval [1, 3, 5, 10, 15, 20]. The results are shown in Fig.~\ref{fig:k}. 
Overall, our method outperforms StandardRAG across different values of $K$.
Moreover, our method achieves stronger performance than StandardRAG when given less context. We speculate that this is because our model can better utilize internal knowledge, especially when given less information. 
However, our advantage decreases when the number of documents is too large. 
We speculate that this is due to the long context bottleneck of the model.

In summary, our method demonstrates strong robustness across different context sizes and provides a practical solution in settings with limited external information or constrained retrieval capacity.

\subsection{Inference Efficiency}
To evaluate the prospect of CoCoA in practical applications, we study its inference efficiency. 
As shown in Table~\ref{tab:efficiency_2wikimqa}, our CoCoA is three times more expensive than common methods (eg., CON), requiring a budget of about 600 output tokens. However, compared to reasoning models (eg., R1), the cost still has a significant advantage, especially when the performance is improved. This achieves a trade-off between performance and efficiency.
\begin{table}[!h]
\centering
\resizebox{0.95\linewidth}{!}{%
\begin{tabular}{lccc}
\toprule
Method & Avg. Input & Avg. Output & Avg(EM,F1) \\
\midrule
CoT     & 904.184 & 195.882 & 23.83\\
R1-8B & 829.186 & 818.200  & 31.30 \\
CoCoA   & 993.184 & 609.742 & \textbf{41.29}\\
\bottomrule
\end{tabular}
}
\caption{Efficiency analysis on 2WikiMQA, with average token count reported.}
\label{tab:efficiency_2wikimqa}
\end{table}

\section{Conclusion}
We investigate the challenge of parametric–retrieved knowledge synergy and introduce \textbf{CoCoA}, a RAG framework that improves LLM performance. 
By leveraging a two-stage multi-agent pipeline, CoCoA-zero 
integrates internal and external knowledge and provides self-synthesized supervisory signals.
With long-chain training, CoCoA delivers strong results on QA tasks, demonstrating its effectiveness and offering insights into long-chain reasoning and collaborative agent training for knowledge-intensive applications.

\clearpage
\newpage
\section*{Limitations}
While CoCoA has demonstrated excellent performance and provided valuable insights into collaboration with parametric and retrieved knowledge, there are still some limitations: 
\begin{itemize}
    \item The current design focuses on a specific agent collaboration pattern via long-chain training. Its applicability to broader or alternative multi-agent architectures remains to be examined. 
    \item Although the approach performs robustly under limited supervision, its scaling with respect to larger models and datasets has not been systematically explored. 
    \item Although the performance has been improved, the token consumption has increased, which has certain limitations in practical applications. How to accelerate reasoning is still a future research direction.
\end{itemize} 

\bibliography{custom}

\begin{thebibliography}{40}
\providecommand{\natexlab}[1]{#1}

\bibitem[{Achiam et~al.(2023)Achiam, Adler, Agarwal, Ahmad, Akkaya, Aleman, Almeida, Altenschmidt, Altman, Anadkat et~al.}]{achiam2023gpt}
Josh Achiam, Steven Adler, Sandhini Agarwal, Lama Ahmad, Ilge Akkaya, Florencia~Leoni Aleman, Diogo Almeida, Janko Altenschmidt, Sam Altman, Shyamal Anadkat, and 1 others. 2023.
\newblock Gpt-4 technical report.
\newblock \emph{arXiv preprint arXiv:2303.08774}.

\bibitem[{Asai et~al.(2023)Asai, Wu, Wang, Sil, and Hajishirzi}]{asai2023self}
Akari Asai, Zeqiu Wu, Yizhong Wang, Avirup Sil, and Hannaneh Hajishirzi. 2023.
\newblock Self-rag: Learning to retrieve, generate, and critique through self-reflection.
\newblock \emph{arXiv preprint arXiv:2310.11511}.

\bibitem[{Berant et~al.(2013)Berant, Chou, Frostig, and Liang}]{berant2013semantic}
Jonathan Berant, Andrew Chou, Roy Frostig, and Percy Liang. 2013.
\newblock Semantic parsing on freebase from question-answer pairs.
\newblock In \emph{Proceedings of the 2013 conference on empirical methods in natural language processing}, pages 1533--1544.

\bibitem[{Cheng et~al.(2024)Cheng, Li, Li, Zhu, Yin, Shao, Li, Sun, Yan, and Qiu}]{cheng2024unified}
Qinyuan Cheng, Xiaonan Li, Shimin Li, Qin Zhu, Zhangyue Yin, Yunfan Shao, Linyang Li, Tianxiang Sun, Hang Yan, and Xipeng Qiu. 2024.
\newblock Unified active retrieval for retrieval augmented generation.
\newblock \emph{arXiv preprint arXiv:2406.12534}.

\bibitem[{Fan et~al.(2024)Fan, Ding, Ning, Wang, Li, Yin, Chua, and Li}]{fan2024survey}
Wenqi Fan, Yujuan Ding, Liangbo Ning, Shijie Wang, Hengyun Li, Dawei Yin, Tat-Seng Chua, and Qing Li. 2024.
\newblock A survey on rag meeting llms: Towards retrieval-augmented large language models.
\newblock In \emph{Proceedings of the 30th ACM SIGKDD Conference on Knowledge Discovery and Data Mining}, pages 6491--6501.

\bibitem[{Gao et~al.(2023)Gao, Xiong, Gao, Jia, Pan, Bi, Dai, Sun, and Wang}]{gao2023retrieval}
Yunfan Gao, Yun Xiong, Xinyu Gao, Kangxiang Jia, Jinliu Pan, Yuxi Bi, Yi~Dai, Jiawei Sun, and Haofen Wang. 2023.
\newblock Retrieval-augmented generation for large language models: A survey.
\newblock \emph{arXiv preprint arXiv:2312.10997}.

\bibitem[{Glass et~al.(2022)Glass, Rossiello, Chowdhury, Naik, Cai, and Gliozzo}]{glass2022re2g}
Michael Glass, Gaetano Rossiello, Md~Faisal~Mahbub Chowdhury, Ankita Naik, Pengshan Cai, and Alfio Gliozzo. 2022.
\newblock Re2g: Retrieve, rerank, generate.
\newblock In \emph{Proceedings of the 2022 Conference of the North American Chapter of the Association for Computational Linguistics: Human Language Technologies}, pages 2701--2715.

\bibitem[{Guo et~al.(2025)Guo, Yang, Zhang, Song, Zhang, Xu, Zhu, Ma, Wang, Bi et~al.}]{guo2025deepseek}
Daya Guo, Dejian Yang, Haowei Zhang, Junxiao Song, Ruoyu Zhang, Runxin Xu, Qihao Zhu, Shirong Ma, Peiyi Wang, Xiao Bi, and 1 others. 2025.
\newblock Deepseek-r1: Incentivizing reasoning capability in llms via reinforcement learning.
\newblock \emph{arXiv preprint arXiv:2501.12948}.

\bibitem[{He et~al.(2022)He, Zhang, and Roth}]{he2022rethinking}
Hangfeng He, Hongming Zhang, and Dan Roth. 2022.
\newblock Rethinking with retrieval: Faithful large language model inference.
\newblock \emph{arXiv preprint arXiv:2301.00303}.

\bibitem[{Ho et~al.(2020{\natexlab{a}})Ho, Nguyen, Sugawara, and Aizawa}]{yang2018hotpotqa}
Xanh Ho, Anh-Khoa~Duong Nguyen, Saku Sugawara, and Akiko Aizawa. 2020{\natexlab{a}}.
\newblock Constructing a multi-hop qa dataset for comprehensive evaluation of reasoning steps.
\newblock \emph{arXiv preprint arXiv:2011.01060}.

\bibitem[{Ho et~al.(2020{\natexlab{b}})Ho, Nguyen, Sugawara, and Aizawa}]{ho2020constructing}
Xanh Ho, Anh-Khoa~Duong Nguyen, Saku Sugawara, and Akiko Aizawa. 2020{\natexlab{b}}.
\newblock Constructing a multi-hop qa dataset for comprehensive evaluation of reasoning steps.
\newblock \emph{arXiv preprint arXiv:2011.01060}.

\bibitem[{Izacard et~al.(2021)Izacard, Caron, Hosseini, Riedel, Bojanowski, Joulin, and Grave}]{izacard2021unsupervised}
Gautier Izacard, Mathilde Caron, Lucas Hosseini, Sebastian Riedel, Piotr Bojanowski, Armand Joulin, and Edouard Grave. 2021.
\newblock Unsupervised dense information retrieval with contrastive learning.
\newblock \emph{arXiv preprint arXiv:2112.09118}.

\bibitem[{Ji et~al.(2023)Ji, Lee, Frieske, Yu, Su, Xu, Ishii, Bang, Madotto, and Fung}]{ji2023survey}
Ziwei Ji, Nayeon Lee, Rita Frieske, Tiezheng Yu, Dan Su, Yan Xu, Etsuko Ishii, Ye~Jin Bang, Andrea Madotto, and Pascale Fung. 2023.
\newblock Survey of hallucination in natural language generation.
\newblock \emph{ACM Computing Surveys}, 55(12):1--38.

\bibitem[{Joshi et~al.(2017)Joshi, Choi, Weld, and Zettlemoyer}]{joshi2017triviaqa}
Mandar Joshi, Eunsol Choi, Daniel~S Weld, and Luke Zettlemoyer. 2017.
\newblock Triviaqa: A large scale distantly supervised challenge dataset for reading comprehension.
\newblock In \emph{Proceedings of the 55th Annual Meeting of the Association for Computational Linguistics (Volume 1: Long Papers)}, pages 1601--1611.

\bibitem[{Karpukhin et~al.(2020)Karpukhin, O{\u{g}}uz, Min, Lewis, Wu, Edunov, Chen, and Yih}]{karpukhin2020dense}
Vladimir Karpukhin, Barlas O{\u{g}}uz, Sewon Min, Patrick Lewis, Ledell Wu, Sergey Edunov, Danqi Chen, and Wen-tau Yih. 2020.
\newblock Dense passage retrieval for open-domain question answering.
\newblock \emph{arXiv preprint arXiv:2004.04906}.

\bibitem[{Kim et~al.(2024{\natexlab{a}})Kim, Nam, Mo, Park, Lee, Seo, Ha, and Shin}]{kim2024sure}
Jaehyung Kim, Jaehyun Nam, Sangwoo Mo, Jongjin Park, Sang-Woo Lee, Minjoon Seo, Jung-Woo Ha, and Jinwoo Shin. 2024{\natexlab{a}}.
\newblock Sure: Summarizing retrievals using answer candidates for open-domain qa of llms.
\newblock \emph{arXiv preprint arXiv:2404.13081}.

\bibitem[{Kim and Lee(2024)}]{kim2024re}
Kiseung Kim and Jay-Yoon Lee. 2024.
\newblock Re-rag: Improving open-domain qa performance and interpretability with relevance estimator in retrieval-augmented generation.
\newblock In \emph{Proceedings of the 2024 Conference on Empirical Methods in Natural Language Processing}, pages 22149--22161.

\bibitem[{Kim et~al.(2024{\natexlab{b}})Kim, Kim, Park, Park, Cho, Kim, Yoo, Lee, and Kim}]{kim2024adaptive}
Youna Kim, Hyuhng~Joon Kim, Cheonbok Park, Choonghyun Park, Hyunsoo Cho, Junyeob Kim, Kang~Min Yoo, Sang-goo Lee, and Taeuk Kim. 2024{\natexlab{b}}.
\newblock Adaptive contrastive decoding in retrieval-augmented generation for handling noisy contexts.
\newblock In \emph{Findings of the Association for Computational Linguistics: EMNLP 2024}, pages 2421--2431.

\bibitem[{Kwon et~al.(2023)Kwon, Li, Zhuang, Sheng, Zheng, Yu, Gonzalez, Zhang, and Stoica}]{kwon2023efficient}
Woosuk Kwon, Zhuohan Li, Siyuan Zhuang, Ying Sheng, Lianmin Zheng, Cody~Hao Yu, Joseph~E. Gonzalez, Hao Zhang, and Ion Stoica. 2023.
\newblock Efficient memory management for large language model serving with pagedattention.
\newblock In \emph{Proceedings of the ACM SIGOPS 29th Symposium on Operating Systems Principles}.

\bibitem[{Lewis et~al.(2020)Lewis, Perez, Piktus, Petroni, Karpukhin, Goyal, K{\"u}ttler, Lewis, Yih, Rockt{\"a}schel et~al.}]{lewis2020retrieval}
Patrick Lewis, Ethan Perez, Aleksandra Piktus, Fabio Petroni, Vladimir Karpukhin, Naman Goyal, Heinrich K{\"u}ttler, Mike Lewis, Wen-tau Yih, Tim Rockt{\"a}schel, and 1 others. 2020.
\newblock Retrieval-augmented generation for knowledge-intensive nlp tasks.
\newblock \emph{Advances in Neural Information Processing Systems}, 33:9459--9474.

\bibitem[{Ma et~al.(2023)Ma, Gong, He, Zhao, and Duan}]{ma2023query}
Xinbei Ma, Yeyun Gong, Pengcheng He, Hai Zhao, and Nan Duan. 2023.
\newblock Query rewriting in retrieval-augmented large language models.
\newblock In \emph{Proceedings of the 2023 Conference on Empirical Methods in Natural Language Processing}, pages 5303--5315.

\bibitem[{Mallen et~al.(2022)Mallen, Asai, Zhong, Das, Hajishirzi, and Khashabi}]{mallen2022not}
Alex Mallen, Akari Asai, Victor Zhong, Rajarshi Das, Hannaneh Hajishirzi, and Daniel Khashabi. 2022.
\newblock When not to trust language models: Investigating effectiveness and limitations of parametric and non-parametric memories.
\newblock \emph{arXiv preprint arXiv:2212.10511}, 7.

\bibitem[{Nian et~al.(2024)Nian, Peng, Wang, and Fang}]{nian2024w}
Jinming Nian, Zhiyuan Peng, Qifan Wang, and Yi~Fang. 2024.
\newblock W-rag: Weakly supervised dense retrieval in rag for open-domain question answering.
\newblock \emph{arXiv preprint arXiv:2408.08444}.

\bibitem[{Rafailov et~al.(2023)Rafailov, Sharma, Mitchell, Manning, Ermon, and Finn}]{rafailov2023direct}
Rafael Rafailov, Archit Sharma, Eric Mitchell, Christopher~D Manning, Stefano Ermon, and Chelsea Finn. 2023.
\newblock Direct preference optimization: Your language model is secretly a reward model.
\newblock \emph{Advances in Neural Information Processing Systems}, 36:53728--53741.

\bibitem[{Shi et~al.(2023)Shi, Han, Lewis, Tsvetkov, Zettlemoyer, and Yih}]{shi2023trusting}
Weijia Shi, Xiaochuang Han, Mike Lewis, Yulia Tsvetkov, Luke Zettlemoyer, and Scott Wen-tau Yih. 2023.
\newblock Trusting your evidence: Hallucinate less with context-aware decoding.
\newblock \emph{arXiv preprint arXiv:2305.14739}.

\bibitem[{Tan et~al.(2024)Tan, Sun, Yang, Wang, Cao, and Cheng}]{tan-etal-2024-blinded}
Hexiang Tan, Fei Sun, Wanli Yang, Yuanzhuo Wang, Qi~Cao, and Xueqi Cheng. 2024.
\newblock Blinded by generated contexts: How language models merge generated and retrieved contexts when knowledge conflicts?
\newblock In \emph{Proceedings of the 62nd Annual Meeting of the Association for Computational Linguistics (Volume 1: Long Papers)}, pages 6207--6227, Bangkok, Thailand. Association for Computational Linguistics.

\bibitem[{Touvron et~al.(2023)Touvron, Lavril, Izacard, Martinet, Lachaux, Lacroix, Rozi{\`e}re, Goyal, Hambro, Azhar et~al.}]{touvron2023LLaMA}
Hugo Touvron, Thibaut Lavril, Gautier Izacard, Xavier Martinet, Marie-Anne Lachaux, Timoth{\'e}e Lacroix, Baptiste Rozi{\`e}re, Naman Goyal, Eric Hambro, Faisal Azhar, and 1 others. 2023.
\newblock Llama: Open and efficient foundation language models.
\newblock \emph{arXiv preprint arXiv:2302.13971}.

\bibitem[{Trivedi et~al.(2022)Trivedi, Balasubramanian, Khot, and Sabharwal}]{trivedi2022interleaving}
Harsh Trivedi, Niranjan Balasubramanian, Tushar Khot, and Ashish Sabharwal. 2022.
\newblock Interleaving retrieval with chain-of-thought reasoning for knowledge-intensive multi-step questions.
\newblock \emph{arXiv preprint arXiv:2212.10509}.

\bibitem[{Wang et~al.(2024{\natexlab{a}})Wang, Wan, Sun, Chen, and Ar{\i}k}]{wang2024astute}
Fei Wang, Xingchen Wan, Ruoxi Sun, Jiefeng Chen, and Sercan~{\"O} Ar{\i}k. 2024{\natexlab{a}}.
\newblock Astute rag: Overcoming imperfect retrieval augmentation and knowledge conflicts for large language models.
\newblock \emph{arXiv preprint arXiv:2410.07176}.

\bibitem[{Wang et~al.(2024{\natexlab{b}})Wang, Li, Jiang, Tian, Wang, Luo, Tang, Cheng, Zhao, and Gao}]{wang2024blendfilter}
Haoyu Wang, Ruirui Li, Haoming Jiang, Jinjin Tian, Zhengyang Wang, Chen Luo, Xianfeng Tang, Monica Cheng, Tuo Zhao, and Jing Gao. 2024{\natexlab{b}}.
\newblock Blendfilter: Advancing retrieval-augmented large language models via query generation blending and knowledge filtering.
\newblock \emph{arXiv preprint arXiv:2402.11129}.

\bibitem[{Wang et~al.(2023{\natexlab{a}})Wang, Yang, and Wei}]{wang2023query2doc}
Liang Wang, Nan Yang, and Furu Wei. 2023{\natexlab{a}}.
\newblock Query2doc: Query expansion with large language models.
\newblock In \emph{Proceedings of the 2023 Conference on Empirical Methods in Natural Language Processing}, pages 9414--9423.

\bibitem[{Wang et~al.(2023{\natexlab{b}})Wang, Li, Sun, and Liu}]{wang2023self}
Yile Wang, Peng Li, Maosong Sun, and Yang Liu. 2023{\natexlab{b}}.
\newblock Self-knowledge guided retrieval augmentation for large language models.
\newblock In \emph{Findings of the Association for Computational Linguistics: EMNLP 2023}, pages 10303--10315.

\bibitem[{Wang et~al.(2024{\natexlab{c}})Wang, Ren, Li, Zhao, Liu, and Wen}]{wang2024rear}
Yuhao Wang, Ruiyang Ren, Junyi Li, Wayne~Xin Zhao, Jing Liu, and Ji-Rong Wen. 2024{\natexlab{c}}.
\newblock Rear: A relevance-aware retrieval-augmented framework for open-domain question answering.
\newblock \emph{arXiv preprint arXiv:2402.17497}.

\bibitem[{Wei et~al.(2022)Wei, Wang, Schuurmans, Bosma, Xia, Chi, Le, Zhou et~al.}]{wei2022chain}
Jason Wei, Xuezhi Wang, Dale Schuurmans, Maarten Bosma, Fei Xia, Ed~Chi, Quoc~V Le, Denny Zhou, and 1 others. 2022.
\newblock Chain-of-thought prompting elicits reasoning in large language models.
\newblock \emph{Advances in neural information processing systems}, 35:24824--24837.

\bibitem[{Wei et~al.(2024)Wei, Chen, and Meng}]{wei2024instructrag}
Zhepei Wei, Wei-Lin Chen, and Yu~Meng. 2024.
\newblock Instructrag: Instructing retrieval-augmented generation via self-synthesized rationales.
\newblock \emph{arXiv preprint arXiv:2406.13629}.

\bibitem[{Xia et~al.(2025)Xia, Zhou, Shi, Chen, and Huang}]{xia2025improving}
Yuan Xia, Jingbo Zhou, Zhenhui Shi, Jun Chen, and Haifeng Huang. 2025.
\newblock Improving retrieval augmented language model with self-reasoning.
\newblock In \emph{Proceedings of the AAAI conference on artificial intelligence}, volume~39, pages 25534--25542.

\bibitem[{Yu et~al.(2022)Yu, Iter, Wang, Xu, Ju, Sanyal, Zhu, Zeng, and Jiang}]{yu2022generate}
Wenhao Yu, Dan Iter, Shuohang Wang, Yichong Xu, Mingxuan Ju, Soumya Sanyal, Chenguang Zhu, Michael Zeng, and Meng Jiang. 2022.
\newblock Generate rather than retrieve: Large language models are strong context generators.
\newblock \emph{arXiv preprint arXiv:2209.10063}.

\bibitem[{Yu et~al.(2023)Yu, Zhang, Pan, Ma, Wang, and Yu}]{yu2023chain}
Wenhao Yu, Hongming Zhang, Xiaoman Pan, Kaixin Ma, Hongwei Wang, and Dong Yu. 2023.
\newblock Chain-of-note: Enhancing robustness in retrieval-augmented language models.
\newblock \emph{arXiv preprint arXiv:2311.09210}.

\bibitem[{Yue et~al.(2025)Yue, Wang, Chen, Huang, and Wei}]{yue2025synergistic}
Shengbin Yue, Siyuan Wang, Wei Chen, Xuanjing Huang, and Zhongyu Wei. 2025.
\newblock Synergistic multi-agent framework with trajectory learning for knowledge-intensive tasks.
\newblock In \emph{Proceedings of the AAAI Conference on Artificial Intelligence}, volume~39, pages 25796--25804.

\bibitem[{Zhang et~al.(2024)Zhang, Patil, Jain, Shen, Zaharia, Stoica, and Gonzalez}]{zhang2024raft}
Tianjun Zhang, Shishir~G Patil, Naman Jain, Sheng Shen, Matei Zaharia, Ion Stoica, and Joseph~E Gonzalez. 2024.
\newblock Raft: Adapting language model to domain specific rag.
\newblock \emph{arXiv preprint arXiv:2403.10131}.

\end{thebibliography}
\appendix

\section{Dataset}
\label{sec:appendix-dataset}
Here, we introduce in detail the datasets we used, which are four datasets on four tasks.

\textbf{2WikiMultiHopQA}~\citep{ho2020constructing} and \textbf{HotpotQA}~\citep{yang2018hotpotqa}: Both datasets are multi-hop question answering datasets based on Wikipedia. Considering the limitation of experimental cost, we used the sub-sampling set published by \citet{trivedi2022interleaving,kim2024sure}, which is obtained by extracting 500 questions from the validation set of each dataset.

\textbf{WebQuestions}~\citep{berant2013semantic}: Constructed from questions posed by the Google Suggest API, where the answers are specific entities listed in Freebase.

\textbf{TriviaQA}~\citep{joshi2017triviaqa}: A compilation of trivia questions paired with answers, both originally pulled from online sources.

\paragraph{Training Data} We sampled subsets from the training sets of HotpotQA~\citep{yang2018hotpotqa}, 2WikiMultiHopQA~\citep{ho2020constructing} and WebQuestions~\citep{berant2013semantic}, then used the CoCoA-zero framework to synthesize data and filtered them with gold answers.
Finally, we selected 6.8k filtered samples, including 3k, 3k, and 0.8k from the three datasets, respectively. 
For the DPO training data, we screen out 1151 samples, which are the ones that are answered incorrectly by zero-shot but correctly by the CoCoA-zero. 
For each sample, we gathered 5 relevant passages using the most common retriever Contriever~\citep{izacard2021unsupervised}.

\begin{table}[!ht]
\centering
\resizebox{0.925\linewidth}{!}{%
\begin{tabular}{@{}ccc@{}}
\toprule
Task Type             & Datasets      & \# Samples \\ 
\hline
\multirow{3}{*}{Multi-HopQA} & 2WikiMultiHopQA & 500    \\
                             & HotpotQA        & 500    \\
\midrule
\multirow{3}{*}{OpenQA}      & WebQuestions    & 2032   \\
                             & TriviaQA       & 11313   \\
\bottomrule
\end{tabular}
}
\caption{Description of tasks and evaluation datasets.}
\label{tab:datasets}
\end{table}

\section{Baselines}
\label{appendix:basleine}
We selected several of the most representative methods for comparison. 
\begin{itemize}
    \item \textbf{StandardRAG}, which is the most classic ``retrieve-then-read’’ paradigm. 
    \item \textbf{Chain-Of-Thought}~\citep{wei2022chain}: Uses CoT prompting to generate reasoning steps before producing the final answer. 
    \item \textbf{Chain-Of-Note}~\citep{yu2023chain}: Refines the retrieved passages prior to answering. 
    \item \textbf{GenRead}~\citep{yu2022generate}: Generates self-contained intermediate context to answer, effectively replacing retrieval with generation. 
    \item \textbf{SURE}~\citep{kim2024sure}: Conditional summarization followed by multiple validation. 
    \item \textbf{Self-RAG}~\citep{asai2023self}: Employs adaptive retrieval and self-reflection to decide when and how to use external knowledge. 
    \item \textbf{DeepSeek-R1-Distill-8B}~\citep{guo2025deepseek}: A distilled LLaMA-8B model released by DeepSeek-R1, trained on reasoning data. 
    \item \textbf{InstructRAG}~\citep{wei2024instructrag}: Denoising training using self-synthesized data. 
\end{itemize}

\paragraph{Baseline Setting.} We followed the original settings for almost all experiments. All retrieval-based methods use top-5 passages. For baselines requiring training, we directly used their weights. Note that InstructRAG directly generates long  rationales, the first half of which consists mostly of analysis and citations of the document, resulting in a non-strictly high EM score and a low F1 score. For a fair comparison, we used Qwen2.5-3B to perform answer segmentation to evaluate. 

\section{Training Details}
\label{sec:appendix-detail}
We fine-tune LLaMA3.1-8B with LoRA (r=16, $\alpha$=16, dropout=0.05) on a maximum input length of 2048. 
LoRA is applied to attention projection layers. 
During SFT, we trained for 5 epochs with a batch size of 1, gradient accumulation of 4, and a learning rate of 3e-5. 
For DPO, a $\beta$ value of 0.2 is applied, using a sigmoid loss function, while RPO is configured with an $\alpha$ value of 0.2. The learning rate was set to 5e-6 and other settings are the same as SFT. 
During inference, we use the vllm~\citep{kwon2023efficient} accelerated inference framework, and to ensure repeatability, we set the temperature to 0.0.
All experiments are conducted on a single A100 GPU with 80GB or 40GB memory. 


\subsection{Performance of Different Model Sizes}
\label{sec:appendix-size}
\begin{figure}[!ht]
    \centering
    \includegraphics[width=0.975\linewidth]{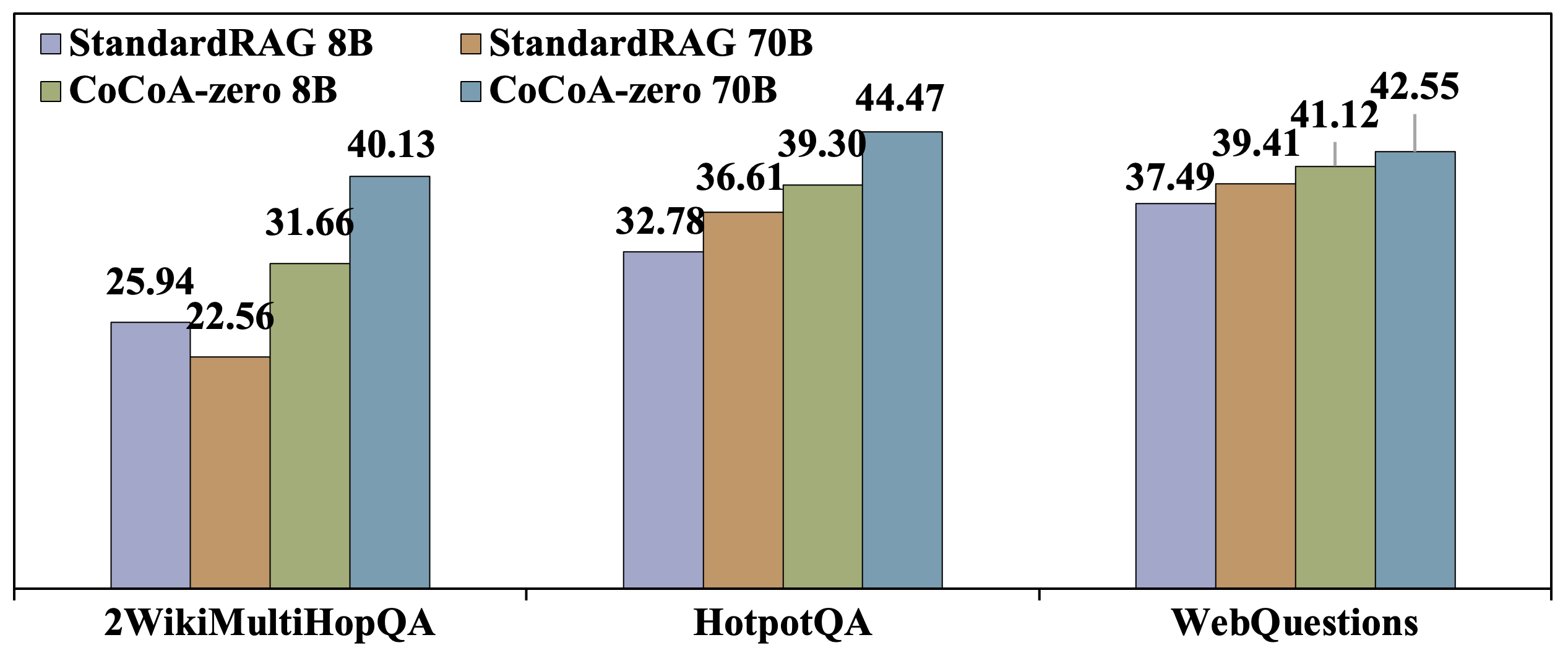}
    \caption{Illustration of performance changes at different model sizes, with Avg(EM,F1) as the metric.}
    \label{fig:model-MS}
\end{figure}

To verify the performance difference of CoCoA-zero under different model sizes, we conducted experiments on performance changes of different model sizes. As shown in Fig.~\ref{fig:model-MS}, the larger the LLM, the better the performance of CoCoA-zero, and it far exceeds standardRAG. 
This shows that larger models better support our collaboration and highlights the importance of internal knowledge in stronger LLMs: the more powerful the LLM, the more it should leverage its internal knowledge for question answering. 

\section{Performance of Different Retriever}
\label{sec:appendix-retriever}
\begin{figure*}[!t]
    \centering
    \includegraphics[width=0.95\linewidth]{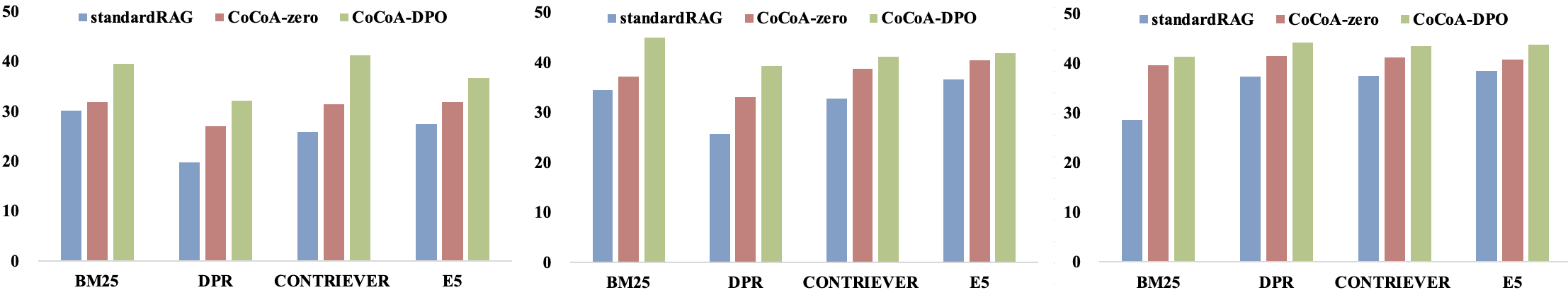}
    \caption{Illustration of performance changes at different retrievers, with Avg(EM,F1) as the metric.}
    \label{fig:retriever}
\end{figure*}

In order to verify the robustness to different retrievers, we selected BM25, DPR, CONTRIEVER and E5 as retrievers. The experimental results are shown in Figure~\ref{fig:retriever}. 
We found that different datasets have different preferences for retrievers, each with its own advantages and disadvantages. 
Overall, both our CoCoA-zero and CoCoA achieve robust performance on different retrievers.

\begin{table*}[!hp]
\centering
\resizebox{0.875\textwidth}{!}{%
\begin{tabular}{lccccccccc}
\toprule
\multirow{2}{*}{Method} & \multicolumn{3}{c}{2WikiMultiHopQA} & \multicolumn{3}{c}{HotpotQA} & \multicolumn{3}{c}{WebQuestions} \\
                        & EM         & F1         & Avg       & EM       & F1      & Avg     & EM        & F1        & Avg      \\
\midrule
CoCoA-zero             & 31.40      & 31.92      & 31.66     & 37.40    & 41.20   & 39.30   & 43.11     & 39.13     & 41.12    \\
-$w/o$ Thinking            & 30.00      & 30.76      & 30.38     & 36.00    & 38.34   & 37.17   & 39.17     & 40.32     & 39.75    \\
-$w/o$ Internal            & 22.60      & 23.93      & 23.26     & 34.00    & 39.11   & 36.56   & 40.01     & 38.20     & 39.10    \\
-$w/o$ External            & 28.40      & 29.53      & 28.97     & 30.00    & 31.92   & 30.96   & 39.81     & 38.13     & 38.97    \\

\midrule

Zero-Shot               & 17.60      & 19.51      & 18.55     & 33.20    & 36.81   & 35.01   & 34.45     & 36.31     & 35.38   \\
Standard RAG     & 26.80      & 25.07      & 25.94     & 31.40    & 34.16   & 32.78   & 37.65     & 37.32     & 37.49    \\
\bottomrule

\end{tabular}%
}
\caption{Ablation study of internal/external induction and reasoning in decision making. In addition, a zero-shot method for explicit internal and external knowledge integration is added for comparison. For simplicity and fairness, the average of EM and F1 is used as the metric.}
\label{tab:appendix-CoA}
\end{table*}

\begin{table*}[hp]
\centering
\resizebox{0.9\textwidth}{!}{%
\begin{tabular}{lccccccccc}
\toprule
\multirow{2}{*}{Method} & \multicolumn{3}{r}{2WikiMultiHopQA} & \multicolumn{3}{r}{HotpotQA} & \multicolumn{3}{r}{WebQuestions} \\
                        & EM         & F1         & Avg       & EM       & F1      & Avg     & EM        & F1        & Avg      \\
\midrule
$\text{Long-DPO}_{8B} $          & 42.00  & 40.58    & 41.29   & 39.00   & 43.39   & 41.20   & 44.83   & 42.21   & 43.52    \\
$\text{Long-SFT}_{8B} $          & 41.00  & 36.87    & 38.94   & 39.40   & 46.31   & 42.86   & 42.96   & 41.32   & 42.14    \\
$\text{Short-SFT}_{8B}$          & 28.60  & 28.03    & 28.31   & 39.00   & 42.15   & 40.58   & 41.19   & 38.48   & 39.84    \\
$\text{Short-SFT}_{8B \times 3}$ & 35.00  & 32.81    & 33.91   & 37.60   & 42.48   & 40.04   & 41.29   & 38.96   & 40.13  \\
\bottomrule
\end{tabular}%
}
\caption{Ablation study of the training strategy for CoCoA. For simplicity and fairness, the average of EM and F1 is used as the metric}
\label{tab:appendix-CoCoA-Train}
\end{table*}

\section{Optimization Analysis} 
\label{sec:appendix-theory}
We analyze the difference between independent training and long chain training in terms of the form of loss. 
We simplify the steps in this analysis, i.e., there are only two steps in the chain, pre-generation processing first and then answering.

When the two agents optimize independently, the loss takes the following form: 
\begin{equation}
\label{eq:pipe-apx}
\begin{aligned}
    \mathcal{L}_{\text{indep}}
   &= -\log P_{\theta}\!\bigl(s \mid x,d\bigr)
      \;-\;\log P_{\phi}\!\bigl(\hat{a} \mid s\bigr).
\end{aligned}
\end{equation}
Here, $\theta$ and $\theta '$ are optimized independently. 

When two agents use long chain optimization, the loss is as follows:
\begin{equation}
\label{eq:long}
\begin{aligned}
\mathcal{L}_{\text{chain}}
    &= -\log P_{\theta}\!\bigl(s,\hat{a} \mid x,d\bigr) \\
    &= -\log P_{\theta}\!\bigl(s \mid x,d\bigr)
      \;-\;\log P_{\theta}\!\bigl(\hat{a} \mid x,d,s\bigr).
\end{aligned}
\end{equation}

\paragraph{Gradient propagation:}~\\ 

The gradient of the first term in Eq.~\eqref{eq:pipe-apx} is,
\begin{equation}
\label{eq:pipe-g}
\begin{aligned}
    \frac{\partial \mathcal{L}_{\text{indep}}}{\partial \theta}
      &= \frac{\partial\,\bigl[-\log P_{\theta}(s \mid x,d)\bigr]}{\partial \theta}
\end{aligned}
\end{equation}

The gradient of the Eq.~\eqref{eq:long} is,
\begin{equation}
\begin{aligned}
    \frac{\partial \mathcal{L}_{\text{chain}}}{\partial \theta}
      &= \frac{\partial\,\bigl[-\log P_{\theta}(s \mid x,d)\bigr]}{\partial \theta}
      \\
      &\qquad  + \frac{\partial\,\bigl[-\log P_{\theta}( \hat{a} \mid x,s,d)\bigr]}{\partial \theta} 
      \\
      &= \eqref{eq:pipe-g}\;+\;\Delta_g
\end{aligned}
\end{equation}

\begin{equation}
\Delta_g \;:=\;
\frac{\partial\,\bigl[-\log P_{\theta}(\hat{a} \mid x,s,d)\bigr]}{\partial \theta}.
\end{equation}
Here, \(\Delta_g\) is the additional gradient that the answer‑loss naturally back‑propagates to the pre-processing parameters when the \emph{same} network \(\theta\) produces both tokens.
In the independent setting \(\Delta_g=0\) by construction, so the preprocessor never “hears” whether the answer is correct, which is not conducive to the consistency of the response. 
The chain objective restores this missing credit assignment signal, thus performing a special kind of multi-task learning on both stages, optimizing them instead of each in isolation, potentially helping to escape from local optimal solutions.

\section{Full Results}
We supplemented the detailed results of the ablation experiment as shown in Table~\ref{tab:appendix-CoA} and Table~\ref{tab:appendix-CoCoA-Train}.



\section{Prompt Templates}
\label{sec:appendix-prompt}
All the prompt templates used by our proposed CoCoA are shown in Table~\ref{tab:prompt-CoCoA-zero} and Table~\ref{tab:prompt-CoCoA}. 
And special instructions are added to section~{3.9} 
corresponding to different tasks as shown in Table~\ref{tab:prompt-task}.

\begin{table*}[!h]
    \centering
    \begin{tabularx}{0.925\linewidth}{lX}
        \toprule
        \textbf{Task}        & \textbf{Task Instruction} \\
        \hline 
        ARC-C       & Given four answer candidates, A, B, C and D, choose the best answer choice. Please answer with the capitalized alphabet only, without adding any extra phrase or period. Do not exceed one word. \\
        \hline
        PubHealth   & Is the following statement correct or not? Say true if it's correct; otherwise say false. Don't capitalize or add periods, just say "true" or "false". Do not exceed one word. \\
        \bottomrule
    \end{tabularx}
    \caption{Full list of instructions used during zero-shot evaluations. For open-domain QA, we don't use any task specific instruction.}
    \label{tab:prompt-task}
\end{table*}

\begin{table*}[h!t]
    \centering
    \begin{tabularx}{0.925\linewidth}{>{\raggedright\arraybackslash}X}
        \toprule
        \textbf{Task}:Prompt used by ``CoCoA’’ \\
        \midrule
        
\#\#\# Instruction:

1. First, provide background for the question. Write a passage that is relevant to the question only based on your knowledge.

2. Second, refer to the provided passages to generate a summary. Cite and write a passage that is relevant to the question only based on the provided passages.

3. Third, refer to the information from the above two sources, verify the accuracy of the facts and the consistency of the logic, and predict the final answer.

\#\#\# Passages:\textbackslash n\{$passages$\}\textbackslash n

\#\#\# Question:\textbackslash n\{$question$\}

\#\#\# Generate Format:

<Internal>\textbackslash nxxx (your background based on your knowledge)\textbackslash n<\textbackslash\textbackslash Internal>

<External>\textbackslash nxxx (your summary based on the provided passages)\textbackslash n<\textbackslash\textbackslash External>

<Thinking>\textbackslash nxxx\textbackslash n<\textbackslash\textbackslash Thinking>

<Answer>\textbackslash nxxx (your short answer consisting of only a few words)<\textbackslash\textbackslash Answer>
        \\
        \bottomrule
    \end{tabularx}
    \caption{The prompt used by ``CoCoA’’.}
    \label{tab:prompt-CoCoA}
\end{table*}

\begin{table*}[h!t]
    \centering
    \begin{tabularx}{\linewidth}{>{\centering\arraybackslash}m{.15\linewidth}X}
        \toprule
        \textbf{Task}        & \textbf{Task Instruction} \\
        \hline
        External Candidate & 
        \#\#\# Passages:{\textbackslash n} \{$passages$\}{\textbackslash n}{\textbackslash n}\newline 
        \#\#\# Instruction:{\textbackslash n} Answer the question below concisely in a few words.{\textbackslash n}{\textbackslash n}\newline
        \#\#\# Input:{\textbackslash n}\{$question$\}{\textbackslash n} \\
        \hline
        External Induction & 
        \#\#\# Instruction:{\textbackslash n} Refer to the provided passages to generate a summary that meets the following conditions:{\textbackslash n} \newline
        1. Cite and Write a passage that can support the prediction about the question only based on the provided passages.{\textbackslash n}\newline
        2. No more than 200 words.{\textbackslash n}\newline
        3. Do not respond with anything other than the \"Summary\".{\textbackslash n}\newline
        \#\#\# Passages:{\textbackslash n} \{$passages$\}{\textbackslash n}{\textbackslash n}\newline
        \#\#\# Question:{\textbackslash n} \{$question$\}{\textbackslash n}\newline
        \#\#\# Prediction:{\textbackslash n} \{$answer$\}{\textbackslash n}{\textbackslash n}\newline
        \#\#\# Generate Format:{\textbackslash n}\newline
        \#\#\# Summary: xxx{\textbackslash n} \\
        \hline
        Internal Candidate & 
        \#\#\# Instruction:{\textbackslash n} Answer the question below concisely in a few words.{\textbackslash n}{\textbackslash n}\newline\#\#\# Input:{\textbackslash n}\{$question$\}{\textbackslash n} \\
        \hline
        Internal Induction       &
        \#\#\# Instruction:{\textbackslash n} Please provide background for the question that meets the following conditions:{\textbackslash n}\newline
        1. Write a passage that can support the prediction about the question only based on your knowledge.{\textbackslash n}\newline
        2. No more than 200 words.{\textbackslash n}\newline
        3. Do not respond with anything other than the \"Background\".{\textbackslash n}\newline
        \#\#\# Question:{\textbackslash n} \{$question$\}{\textbackslash n}\newline
        \#\#\# Prediction:{\textbackslash n} \{$answer$\}{\textbackslash n}{\textbackslash n}\newline
        \#\#\# Generate Format:{\textbackslash n}\newline
        \#\#\# Background: xxx{\textbackslash n}   \\
        \hline
        Decision-Making     & 
        \#\#\# Internal Reasoning Path: {\textbackslash n}\{$induction_{in}$\}{\textbackslash n}{\textbackslash n} 
        \#\#\# Internal Prediction 1: {\textbackslash n}\{$answer_{in}$\}{\textbackslash n}{\textbackslash n}  \newline
        \#\#\# External Reasoning Path: {\textbackslash n}\{$induction_{ex}$\}{\textbackslash n}{\textbackslash n} 
        \#\#\# External Prediction 2: {\textbackslash n}\{$answer_{ex}$\}{\textbackslash n}{\textbackslash n}  \newline
        \#\#\# Instruction:{\textbackslash n} \newline
        Refer to the information from the above two sources, verify the accuracy of the facts and the consistency of the logic, and choose the best prediction. \newline
        \#\#\# Question:{\textbackslash n}\{$question$\}{\textbackslash n}\newline
        \#\#\# Generate Format:{\textbackslash n}\newline
        \#\#\# Thingking: xxx (Please think step by step){\textbackslash n}\newline
        \#\#\# Short Answer: xxx (just in a few words){\textbackslash n} \\
       
        \bottomrule
    \end{tabularx}
    \caption{A list of prompts used by CoCoA-zero.}
    \label{tab:prompt-CoCoA-zero}
\end{table*}

\section{Case Study}
\begin{figure*}[!h]
    \centering
    \includegraphics[width=0.95\linewidth]{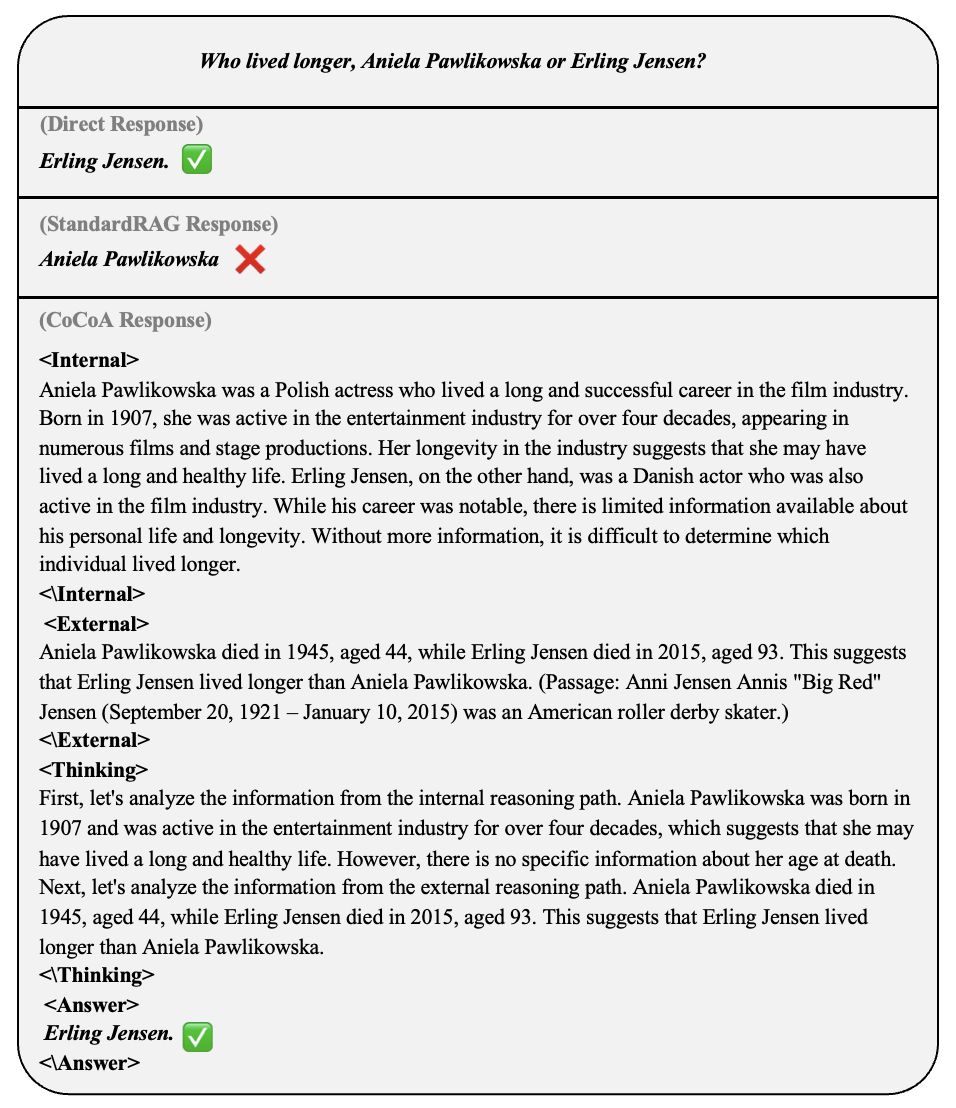}
    \caption{Direct generation is correct, while standard RAG generation is incorrect. This is due to external noise causing RAG to misjudge, while CoCoA performs well in synergy.}
    \label{fig:case1}
\end{figure*}

\begin{figure*}[!h]
    \centering
    \includegraphics[width=0.95\linewidth]{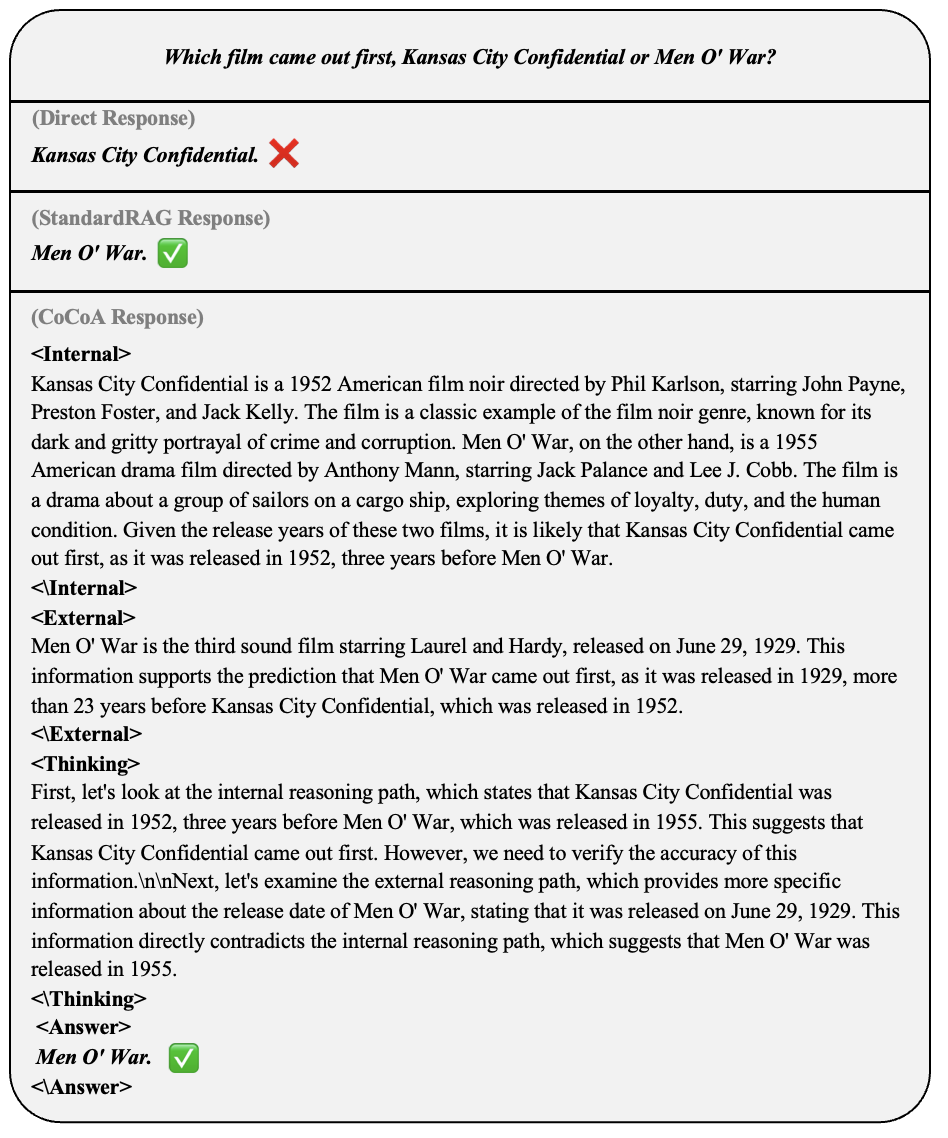}
    \caption{Direct generation is incorrect, while standard RAG is correct. This is due to internal hallucinations causing misjudgment, while CoCoA performs well.}
    \label{fig:case2}
\end{figure*}

\begin{figure*}[!h]
    \centering
    \includegraphics[width=0.95\linewidth]{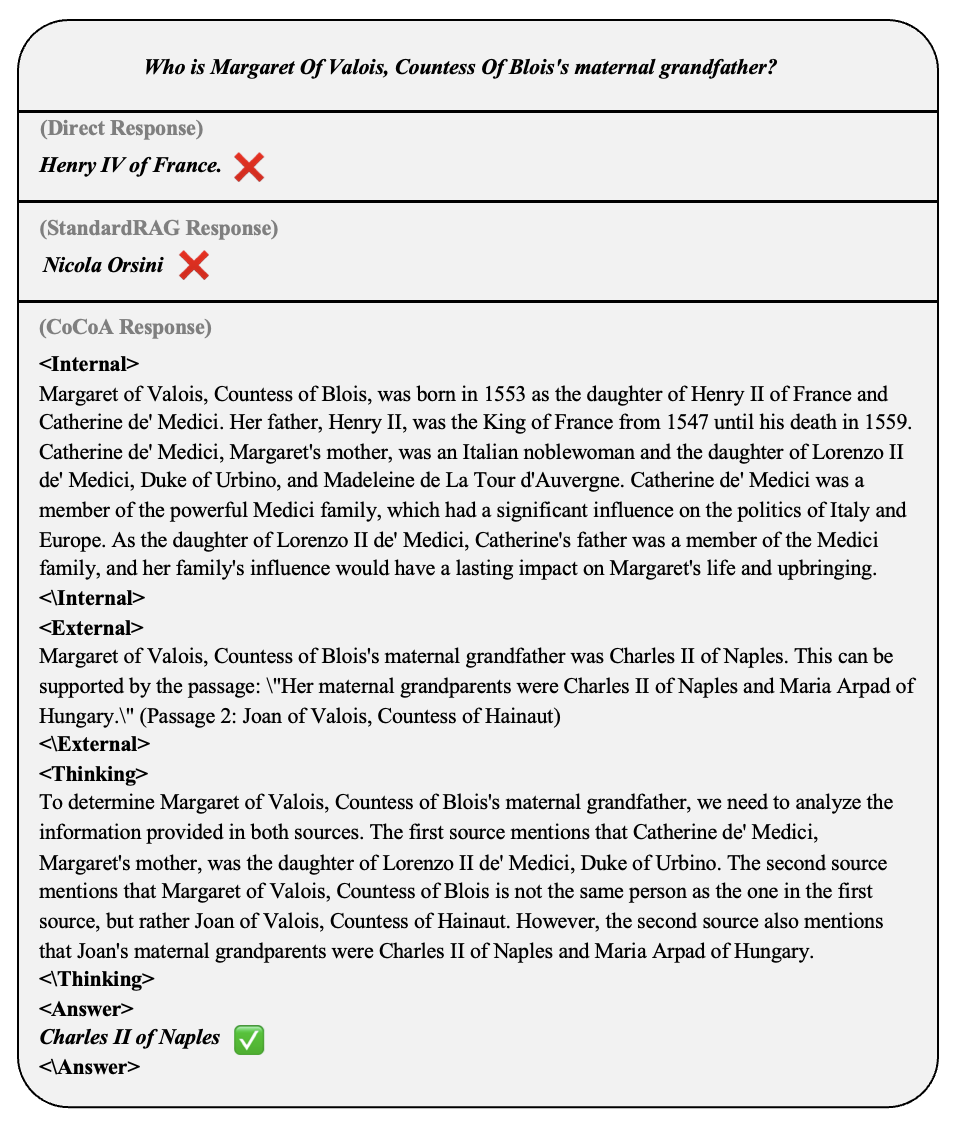}
    \caption{Direct generation is incorrect, and standard RAG generation is also incorrect. CoCoA synergizes well, benefiting from the interaction between knowledge induction and high-level decision-making to stimulate model capabilities.}
    \label{fig:case3}
\end{figure*}

We provide detailed case study in Figure~\ref{fig:case1}, Figure~\ref{fig:case2}, and Figure~\ref{fig:case3}. 


\end{document}